\documentclass[a4paper, 10pt, conference]{ieeeconf}

\IEEEoverridecommandlockouts
\usepackage[export]{adjustbox}
\usepackage{amsmath}
\usepackage{amssymb}
\usepackage[font=small]{caption}
\usepackage{graphicx}
\usepackage[hyphens]{url}

\title{\LARGE \bf Multi-Session Ground Texture SLAM in Low-Dynamic Environments\vspace{-4 mm}}

\author{Kyle M. Hart$^{1,2}$ and Brendan Englot$^{2}$
\thanks{$^{1}$Naval Air Warfare Center, Aircraft Division, Lakehurst, NJ, 08733, USA, {\tt\small kyle.m.hart2.civ@us.navy.mil}}%
\thanks{$^{2}$Department of Mechanical Engineering, Stevens Institute of Technology, Hoboken, NJ, 07030, USA, {\tt\small benglot@stevens.edu}}%
\thanks{Distribution Statement A\@: Approved for public release; distribution is unlimited, as submitted under NAVAIR Public Release Authorization 2025-0098. The views expressed here are those of the authors and do not reflect the official policy or position of the United States Navy, Department of Defense, or United States Government.}
}

\begin{document}

\maketitle
\thispagestyle{empty}
\pagestyle{empty}

\begin{abstract}
The simultaneous localization and mapping community has introduced a growing number of systems adapted for multi-session operations where the operational environment features low-dynamic changes that impact mapping, such as surface wear, weather phenomena, or seasonal change. These systems allow for lifelong operations by a robot within these environments. There is also growing interest in operations in environments where the unique ground texture is the only mapping feature available for use. These ground texture systems are not yet targeted for multi-session low-dynamic-change environments though. This work explores the impact of three different techniques on trajectory estimation accuracy in these multi-session low-dynamic ground texture environments. Of the three, the use of Kullback-Leibler Divergence, as a similarity score and a bias influencing loop closure confidence, is found to have the most success. We show an analysis of all three methods and a deeper exploration of the impact of Kullback-Leibler Divergence. We also introduce a dataset for use by the robotics community that contains multi-session images where the ground changes between sessions and also high-accuracy pose information for use in evaluation.
\end{abstract}
\vspace{-2 mm}

\section{Introduction}
\label{section:introduction}
\vspace{-1 mm}
Monocular visual Simultaneous Localization and Mapping (SLAM) is a common set of techniques often used by ground robot platforms to map and navigate in their environments. These SLAM systems provide accurate, real-time location information using a sensor that is both affordable and information dense. Most visual SLAM systems rely on outward features, such as buildings. Their operating environments often contain numerous such features, making these features natural choices. However, there is a growing volume of work investigating the use of the ground texture beneath the robot as a source of features. This approach is particularly relevant in areas with few outward features, such as large warehouses. 

In order for ground texture based systems to operate over long lifetimes, they must be capable of adapting to low-dynamic changes, such as surface texture wear. Without this capability, a user must manually reinitialize the map periodically as the environment changes. This necessitates additional human operations and limits the effectiveness of the autonomous system.

Here, we develop and investigate multiple methods to improve a ground texture SLAM system's loop closure performance, allowing the system to better handle low-dynamic changes in the environment. To the best of our knowledge, this is the first such investigation for low-dynamic-change ground texture environments. By developing such a method, a robot can operate for longer durations without user intervention, while maintaining high location accuracy. In particular, our contributions are as follows:

\begin{itemize}
    \item A first-of-its-kind analysis of three methods applied to ground texture SLAM for multi-session, low-dynamic-change environments. Namely:
    \begin{itemize}
        \item Use of Kullback-Leibler Divergence as a similarity metric between an observation and a baseline to adjust the confidence in loop closures. This method is found to be the most effective.
        \item Estimating visual overlap between observations using estimated poses and fields of view to exclude unlikely loop closures.
        \item Use of symmetry of joint intensity histograms between pairs of images as a similarity metric to adjust the confidence in loop closures.
    \end{itemize}
    \item The creation of the first ground texture dataset featuring:
    \begin{itemize}
        \item Multiple sessions with low-dynamic changes occurring over large portions of the operating area between sessions.
        \item High-quality ground truth information captured with a motion capture system for use in evaluating accuracy of SLAM systems.
    \end{itemize}
\end{itemize}

In this paper, we will discuss related works in Sec.~\ref{section:related-work}. Then, Sec.~\ref{section:problem-description} defines the multi-session, low-dynamic SLAM problem. Sec.~\ref{section:multi-session-slam-methodologies} describes all the candidate methods. Sec.~\ref{section:evaluation} then presents the dataset used for evaluation, the quantitative results of the evaluation, and some more in-depth observations about the effectiveness of the Kullback-Leibler Divergence approach.

\section{Related Work}
\label{section:related-work}
\vspace{-1 mm}
Visual SLAM is a popular method of robot navigation that uses cameras to perceive the environment, recognize features and landmarks, and build a map for navigation. It is popular in part because cameras are affordable, common, and a rich source of information about the environment. There are numerous approaches in the existing literature. Some, such as Tateno et al.~\cite{tateno_cnn-slam_2017} and Zubizarreta et al.~\cite{zubizarreta_direct_2020}, use direct methods to compare images using pixel intensities. Others use indirect methods that detect and describe keypoints in the image and subsequently use these keypoints for loop closure identification. Of these systems, ORB-SLAM and its successors are some of the most popular~\cite{mur-artal_orb-slam_2015, mur-artal_orb-slam2_2017, campos_orb-slam3_2021}. In most of these systems, the sensors look out and forward into the environment to recognize distinctive features from objects such as furniture, buildings, or vegetation.

However, there is a growing field focused on situations where these types of distinctive external features are not available, such as in warehouses or on empty rural roads. These systems instead rely on the unique ground texture below a robot for localization, mapping, and navigation. Most of this field focuses on localization methods that use a premapped environment to quickly and accurately identify a robot's location within that environment~\cite{schmid_ground_2020, zhang_high-precision_2019, wilhelm_lightweight_2024}. Zhang et al.~\cite{zhang_high-precision_2019} and Wilhelm and Napp~\cite{wilhelm_lightweight_2024} use keypoint detection and a voting scheme to find revisited locations. Zhang and Rusinkiewicz~\cite{zhang_learning_2018} use a neural network for feature detection. Hart et al.~\cite{hart_monocular_2023} extends the field from localization to SLAM. In that work, we use feature detection and description, combined with several thresholds governing loop closure detection. In all cases, these approaches focus on static or approximately static environments. While they may work across multiple sessions, if the ground texture changes significantly, their performance is unknown.

SLAM domains outside the visual ground texture realm explore multi-session scenarios characterized by low-dynamic changes in the environment. Many frameworks focus on detecting landmarks that move between sessions. For example, Walcott-Bryant et al.~\cite{walcott-bryant_dynamic_2012} propose a LIDAR system that handles change due to moving furniture in an indoor environment. Nielsen and Hendeby~\cite{nielsen_feature_2022} maintain multiple hypotheses in their map to adapt to shifting landmarks. Many other systems use submaps and pose graph modification to ensure compact maps. Zhao et al.~\cite{zhao_general_2021} maintain multiple LIDAR-based submaps and use those to update the map as the scene changes. In the visual SLAM domain, Song et al.~\cite{song_g2p-slam_2022} use viewpoint and keypoint constraint consistency checking to discriminate between static, low-dynamic, and high-dynamic features while Lee and Myung~\cite{lee_solution_2014} use error metrics to prune the pose graph created with an RGB-D camera.
\vspace{-4 mm}

\section{Problem Description}
\label{section:problem-description}
\vspace{-2 mm}
We consider the following scenario. The environment consists of an approximately planar ground surface. This ground surface has a single general texture type, such as carpet. The surface is subjected to low-dynamic change in appearance. As described by Walcott-Bryant et al.~\cite{walcott-bryant_dynamic_2012}, this means changes that are not apparent from sequential observations but that are apparent when viewed after sufficient time has passed. Common examples include surface wear or repair. These low-dynamic changes can take place over large sections of the surface such that multiple sequential observations will be impacted by the change. A common example is a repaved road. The surface is not subjected to high-dynamic change, which are rapid changes that may occur between successive frames, such as debris appearing.

Operating in this environment is a robot equipped with a single downward-facing monocular color camera. Following the convention described in Hart et al.~\cite{hart_monocular_2023}, this camera is calibrated with an intrinsic matrix, $\mathbf{K} \in \mathbb{R}^{3 \times 3}$. The camera has a known 3D pose relative to the robot's frame of reference, $R$, as represented by the homogeneous transform matrix $\mathbf{T}_{RC} \in \mathbb{R}^{4 \times 4}$, which transforms measurements from the camera's frame of reference, $C$, to $R$.

The robot moves through the environment during multiple sessions throughout its lifetime. During each session, the robot travels through several poses, $\mathbf{x}_{k,t}$, where $k \in \{0, 1, ..., K\}$ represents the session number and $t$ represents the timestamp of the pose within the session. Note that the total number of poses within each session is not necessarily the same. As this is a ground plane problem, the poses are 2D as measured from the world or map frame, $W$, and can be represented by the homogeneous transform matrix $\mathbf{T}_{W\mathbf{x}_{k,t}} \in \mathbb{R}^{3 \times 3}$, which transforms data in the pose frame of reference to the world frame of reference. At each pose, the robot receives an observation, $\mathbf{z}_{k,t}$, as a distortion-free color image from the downward-facing camera. As the time between the end of one session and start of the next is arbitrarily long, observations from different sessions will observe the low-dynamic change of the ground. There is no requirement to follow the same path, but it is assumed the poses within each session are such that there are a reasonable number of re-observed areas between sessions.

The goal is to find an algorithm that best estimates the robot's poses, $\mathbf{x}_{k,t}$, for all sessions, $k$, and all times, $t$. This algorithm should only use the two camera matrices, $\mathbf{K}$ and $\mathbf{T}_{RC}$; observations from previous sessions, $\mathbf{z}_{0:k-1,0:T}$; and the previous and current observations from the current session, $\mathbf{z}_{k,0:t}$. In other words, the goal is an online SLAM algorithm.
\vspace{-2 mm}

\section{Multi-Session SLAM Methodologies}
\label{section:multi-session-slam-methodologies}
\vspace{-0.5 mm}
As described in Sec.~\ref{section:related-work}, the current set of ground texture localization and SLAM systems are not designed or tested for multi-session scenarios with low-dynamic ground texture changes. As such, there is an opportunity to explore a number of methodologies, inspired by multi-session systems in other domains, to determine which best improves overall system performance. After a brief review of the base SLAM system, each is described briefly below.
\vspace{-1 mm}

\subsection{Foundational System for Ground-Texture Visual SLAM}
\label{section:foundational-system}
\begin{figure}[ht]
    \centering
    \includegraphics[width=1.0\linewidth]{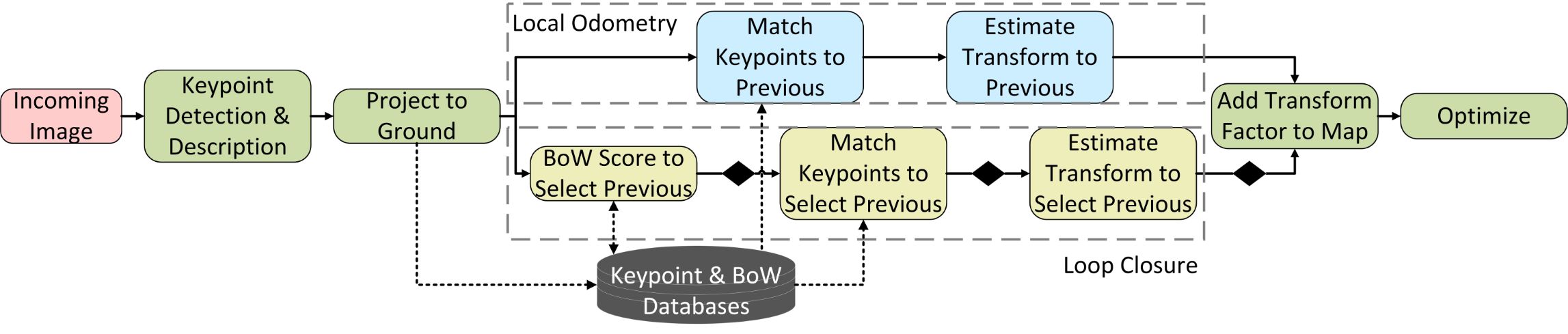}
    \caption{The original ground texture SLAM system introduced by Hart et al.~\cite{hart_monocular_2023}. Our explored methodologies aim to enhance the loop closure process to improve loop closure accuracy in long-term, low-dynamic environments.}
    \label{figure:original-architecture}
    \vspace{-5 mm}
\end{figure}

\vspace{-1 mm}
The proposed methodologies for managing loop closures across SLAM sessions build atop the current state-of-the-art ground texture vision-only SLAM framework proposed by Hart et al.~\cite{hart_monocular_2023}. Fig.~\ref{figure:original-architecture} shows the original architecture of this work upon which all discussed improvements are based. First, incoming images are processed using ORB~\cite{rublee_orb_2011} to identify keypoints and their associated descriptors. These keypoints are then projected onto the ground plane using known information about the camera position. Then, successive pairs of images are used to estimate visual-only odometry. Loop closures are subsequently exploited to correct drift. Both the odometry and loop closure steps use keypoints from the image along with their associated descriptors. Additionally, both steps estimate the transform between pairs of images using keypoints projected into the ground plane and M-estimators, which are robust deterministic models. Unlike the local visual odometry, loop closure detection uses three metrics to determine if a candidate loop closure is valid: visual bag of words scores, number of keypoint matches, and a covariance parameter. The transforms estimated from each step are inserted into a factor graph representing the map.

\subsection{Method 1: Kullback-Leibler Divergence}
\label{section:kullback-leibler-divergence}
\vspace{-1 mm}
The first multi-session approach considered uses Kullback-Leibler Divergence (KLD)~\cite{kullback_information_1951} as a measure of ground texture dissimilarity, which is subsequently used to bias the confidence of loop closure factors added to a factor graph.

KLD is a common statistical measure used to describe the difference between two distributions. It has been used in some SLAM applications, for example by Shiguang and Chengdong~\cite{shiguang_improved_2017}. Here, we calculate the average KLD score for the pixel intensities of each channel of an image and compare to the pixel intensities for each channel of the set of images from the \textit{first session} as a baseline.

During the first session, we store all received images, $\mathbf{z}_{0,0:T}$, creating three normalized histograms of pixel intensity, one per color channel. Assuming 8-bit RGB images, the distributions are as follows:

\vspace{-4 mm}
\begin{equation}
    q_c = \left\{\frac{I_c(\mathbf{z}_{0,0:T}) = 0}{TotalPixelCount}, ..., \frac{I_c(\mathbf{z}_{0,0:T}) = 255}{TotalPixelCount}\right\}
\end{equation}

Here, $I_c$ represents a count of all the pixels from one channel in the observation set where the intensity is equal to the indicated value (e.g., 0, 255, etc.). The histogram, $q_c$, is then normalized. For an RGB image, this produces three distributions, with $c \in \left\{r, g, b\right\}$ representing the three color channels.

In subsequent sessions, as each observation is received, the system then computes a similar set of distributions for just that image. For example, for an 8-bit RGB image, the distributions are as follows:

\vspace{-3 mm}
\begin{equation}
    p_c = \left\{\frac{I_c(\mathbf{z}_{k,t}) = 0}{TotalPixelCount}, ..., \frac{I_c(\mathbf{z}_{k,t}) = 255}{TotalPixelCount}\right\}
\end{equation}

A single KLD score is then computed by calculating the KLD score for each channel and averaging:

\vspace{-2 mm}
\begin{equation}
    KLD_c = \sum_{0}^{255} p_c \log \frac{p_c + \epsilon}{q_c + \epsilon}
\end{equation}

\vspace{-2 mm}
\begin{equation}
    KLD = \frac{(KLD_r + KLD_g + KLD_b)}{3}
\end{equation}

Note that $\epsilon$ is a very small number to prevent division by zero. After obtaining this score, we proceed with the loop closure detection method described above in Sec.~\ref{section:foundational-system}.

However, prior to the last step, that method produces an estimated transform, $\mathbf{T}_{\mathbf{x}_{k,t},\mathbf{x}_{k',t'}}$ and an associated covariance, $\mathbf{\Sigma} \in \mathbb{R}^{3 \times 3}$, that is used for the factor that is potentially added to the pose graph. We modify this covariance as follows and then proceed with the rest of the loop closure estimation method.

\vspace{-4 mm}
\begin{equation}
    \label{equation:kld-covariance-offset}
    \mathbf{\Sigma}^* = \mathbf{\Sigma} \times (KLD + 1)
\end{equation}

In the preceding equation, $\times$ denotes standard scalar multiplication. Additionally, two identical distributions yield a KLD score of $0$, so $1$ is added such that a perfect match yields the multiplicative identity and thus an unchanged covariance, $\mathbf{\Sigma}^* = \mathbf{\Sigma}$.

This is motivated by the use of covariance matrices as measures of confidence within the factor graph framework. Each loop closure has an associated covariance matrix that is used during optimization. Higher variances are weighted less significantly. By scaling the loop closure covariance by the KLD score, loop closure pairs with larger differences (higher KLD scores, lower likelihood of useful matches) end up with higher variances and are thus treated with lower confidence during optimization. This de-emphasizes their impact on the optimized set of pose estimates and avoids false positives.

The same process can also use grayscale images. In this case, there is only one channel, so the average is no longer necessary. The rest of the procedure is the same as described above.
\vspace{-2 mm}

\subsection{Method 2: Visual Overlap}
\label{section:visual-overlap}
\vspace{-1 mm}
Another method considered in our study is the use of predicted field-of-view overlap to eliminate false positives. Inspired by systems that use covisibility consistency, such as the one described in Walcott-Bryant et al.~\cite{walcott-bryant_dynamic_2012}, this approach uses current pose estimates to determine if it is likely for the fields of view of each candidate in the loop closure pair to overlap.

To calculate this, for each estimated pose, we take the coordinates of the image corners and project them onto the robot's frame of reference through the method described by Hart et al.~\cite{hart_monocular_2023}. Then, we use the current estimated robot pose to transform these corners into the map frame, $W$. The result is four points forming a rectangle on the ground plane that represents the estimated field of view. Then, we compute the intersection between the projected rectangle of the current observation and any candidate loop closure observation. Loop closure estimation only proceeds if there is estimated overlap between the two fields of view. This method can also be adjusted to require a minimum overlap area above zero.
\vspace{-2 mm}

\subsection{Method 3: Joint Intensity Histogram Symmetry}
\label{section:joint-intensity-histogram-symmetry}
\vspace{-1 mm}
Similar to KLD, this method seeks to estimate the level of texture change between two images. Joint intensity histograms are often used in change detection between two images, such as the methods described by Kita~\cite{kita_change_2006},~\cite{kita_study_2008}.

This approach begins with the grayscale versions of the images observed at the current pose and at the pose that is being evaluated for potential loop closure. The method computes a 2D histogram of pixel intensities. For 8-bit images, this takes the form shown in Eq. (\ref{equation:jih-matrix}), where $I(i,j)$ is shorthand for the count of pixel locations where in the most recently received image, $\mathbf{z}_{k,t}$, the intensity at a given pixel is a certain value, $I(x, y) = i$, and in the loop closure candidate image, $\mathbf{z}_{k',t'}$, the same pixel locations have a different given value, $I(x, y) = j$.

\vspace{-3 mm}
\begin{equation}
\label{equation:jih-matrix}
    \mathbf{H} = \begin{bmatrix}
        I(0, 0) & \cdots & I(0, 255) \\
        \vdots & \ddots & \vdots \\
        I(255, 0) & \cdots & I(255, 255) \\
    \end{bmatrix}
\end{equation}

Then, the algorithm evaluates the amount of symmetry of the joint intensity histogram using Eqns.~\ref{equation:jih-symmetric}-\ref{equation:jih-score}, which derive from the properties of square matrices. In these steps, $|\cdot|$ represents the Frobenius norm.

\vspace{-3 mm}
\begin{equation}
\label{equation:jih-symmetric}
    \mathbf{H}_{symmetric} = \tfrac{1}{2}(\mathbf{H} + \mathbf{H}^T)
\end{equation}

\vspace{-6 mm}
\begin{equation}
\label{equation:jih-antisymmetric}
    \mathbf{H}_{antisymmetric} = \tfrac{1}{2}(\mathbf{H} - \mathbf{H}^T)
\end{equation}

\vspace{-6 mm}
\begin{equation}
\label{equation:jih-score}
    JIH = \frac{1}{2}\left(\frac{|\mathbf{H}_{symmetric}| - |\mathbf{H}_{antisymmetric}|}{|\mathbf{H}_{symmetric}| + |\mathbf{H}_{antisymmetric}|} + 1\right)
\end{equation}

This approach derives from the observation that similar textures will exhibit similar intensity profiles, resulting in near symmetry for the joint intensity histogram. Inversely, as the texture changes over time, the intensity profile will differ from the original, leading to less symmetry. The result is a score that varies from 0 if the joint intensity histogram is antisymmetric (implying different texture) to 1 if the joint intensity histogram is symmetric (implying similar texture). This can be used to bias the covariance in the loop closure factor, similar to the method described in Sec.~\ref{section:kullback-leibler-divergence}, so that less-symmetric pairs have higher covariance matrices and thus lower confidence and impact on the pose graph.

\vspace{-3 mm}
\begin{equation}
    \mathbf{\Sigma}^* = \mathbf{\Sigma} \times \frac{1}{JIH + \epsilon}
\end{equation}
\vspace{-5 mm}

Note that $\epsilon$ is a very small number in case the joint intensity histogram produces a score of zero. As with Equation (\ref{equation:kld-covariance-offset}), $\times$ is standard scalar multiplication and the terms are structured such that a perfect match leaves the covariance unchanged.
\vspace{-2 mm}

\section{Evaluation}
\label{section:evaluation}
\vspace{-1 mm}
Now that potential methods have been described, the next step is to evaluate their effectiveness in a multi-session, low-dynamic-change environment. This will require a representative dataset and a selection of relevant metrics. After determining the best approach, we present several in-depth results to highlight the impact of the top-performing method.

\vspace{-1 mm}
\subsection{Dataset for Multi-Session Ground-Texture Visual SLAM}
\label{section:dataset}
\vspace{-1 mm}
Because the proposed systems only use observations of the ground texture, it is necessary to use a dataset featuring ground texture images with low-dynamic change. The datasets contributed by Zhang et al.~\cite{zhang_high-precision_2019} and Schmid et al.~\cite{schmid_hd_2022} are the most commonly used in ground texture work. However, both are of static environments. Schmid et al. features a few examples of a wet surface, but neither exhibit large-scale texture changes between sessions. Conversely, there are several visual SLAM datasets featuring low-dynamic change, such as the ones introduced by Gridseth and Barfoot~\cite{gridseth_keeping_2022} and Carlevaris-Bianco et al.~\cite{carlevaris-bianco_university_2016}. However, they feature images looking outward into the environment, rather than at the ground. There is no preexisting dataset that features both elements: ground texture images and low-dynamic changes.

\begin{figure}
    \centering
    \includegraphics[width=0.3\linewidth]{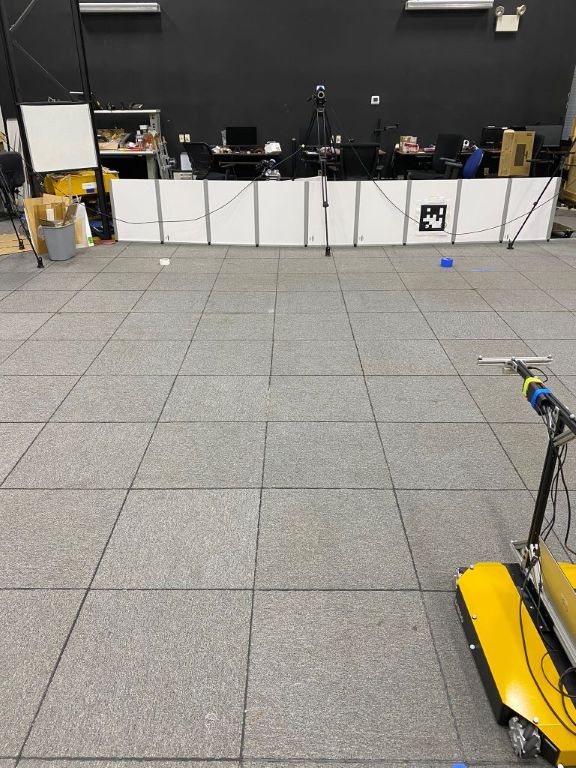}
    \includegraphics[width=0.3\linewidth]{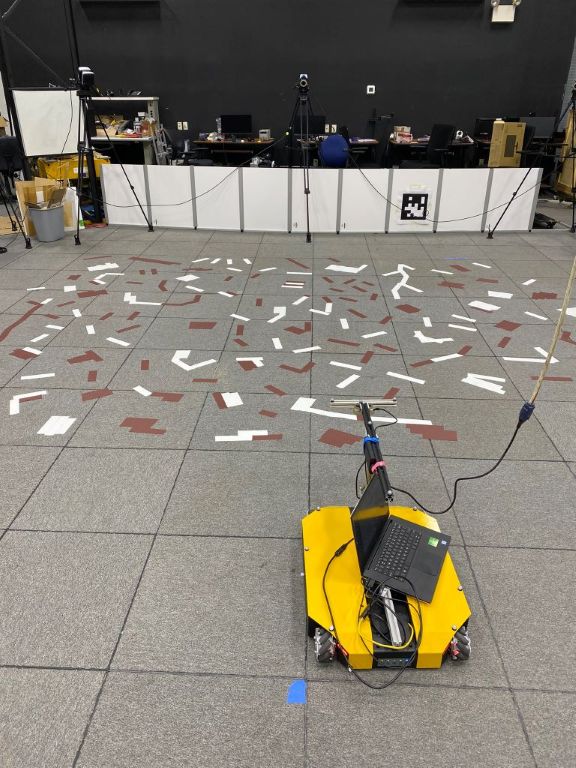}
    \includegraphics[width=0.3\linewidth]{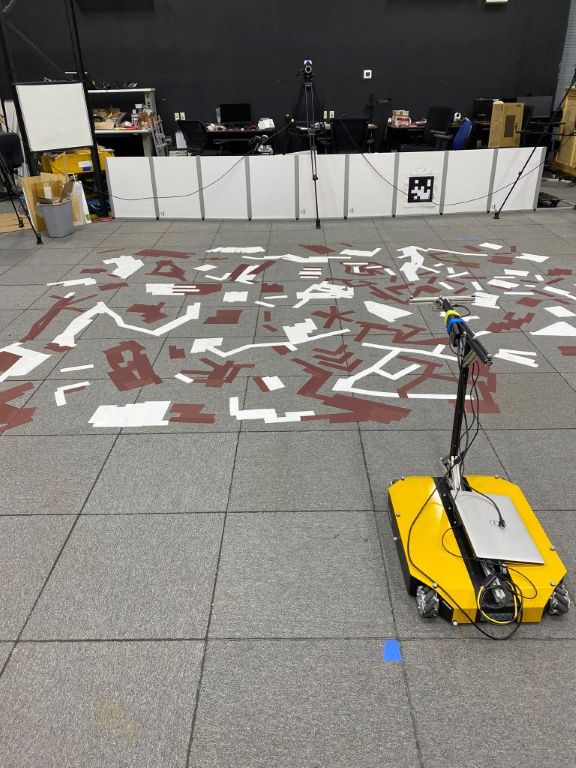}
    \\
    \includegraphics[width=0.3\linewidth]{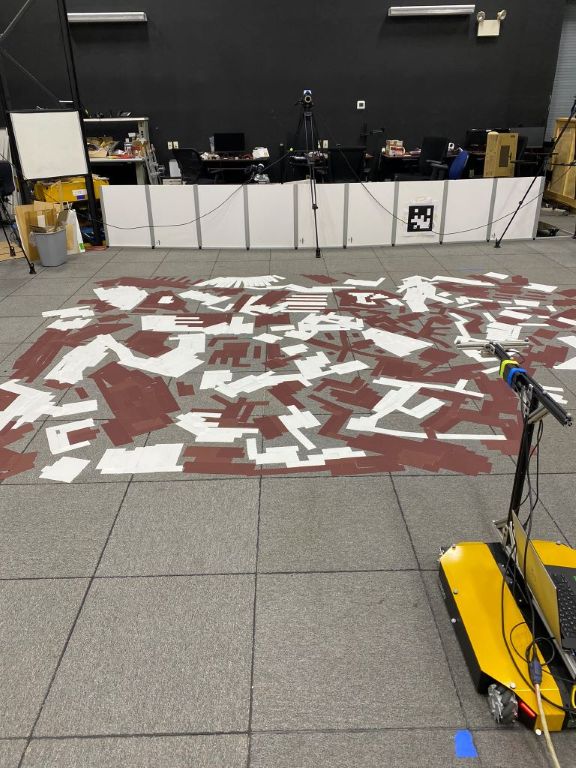}
    \includegraphics[width=0.3\linewidth]{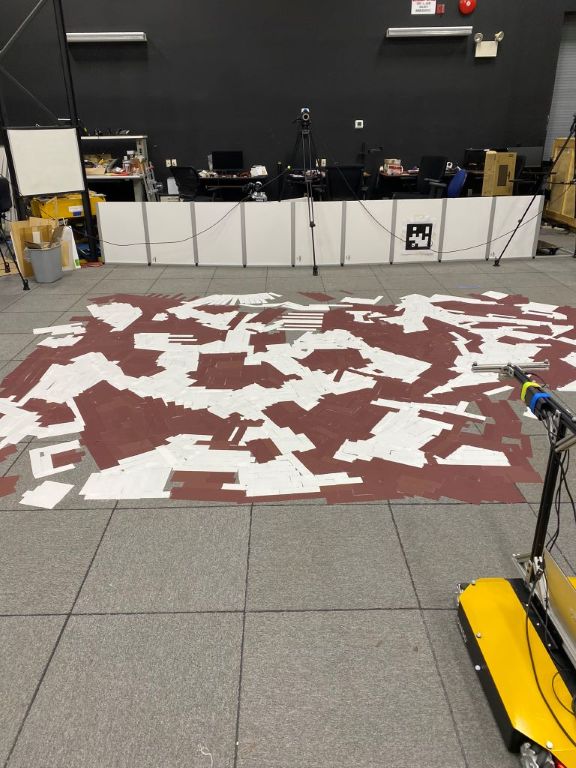}
    \caption{\textbf{Introducing low-dynamic changes in ground texture.} The experiment area, prior to collecting data for each SLAM session.}
    \label{figure:experiment-setup}
    \vspace{-7 mm}
\end{figure}

Therefore, we created and are releasing a ground texture dataset featuring low-dynamic changes between sessions. The environment starts with a carpeted operational area. After each session, tape is placed throughout the environment to simulate wear over time. Each session features more and more tape until almost the entire environment is covered in tape. Fig.~\ref{figure:experiment-setup} shows the setup at the start of each session. We chose this setup to approximate large-scale surface wear, such as rust and flaking of a metal surface or the pitting and appearance change of paved roads over time. As this work is focused on low-dynamic changes, we intentionally excluded high-dynamic changes like debris and lighting changes to isolate the impact of low-dynamic changes.

\begin{figure}
    \centering
    \includegraphics[width=0.45\columnwidth, valign=c]{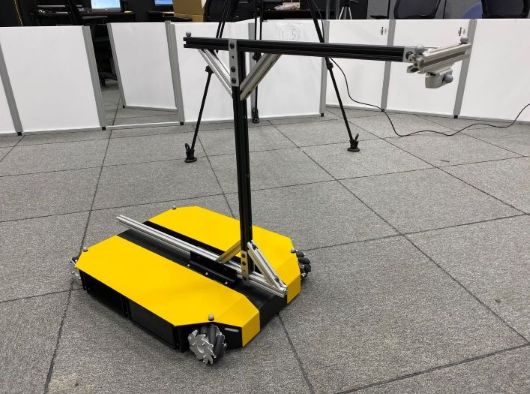}
    \adjustbox{valign=c}{\includegraphics[width=0.45\columnwidth]{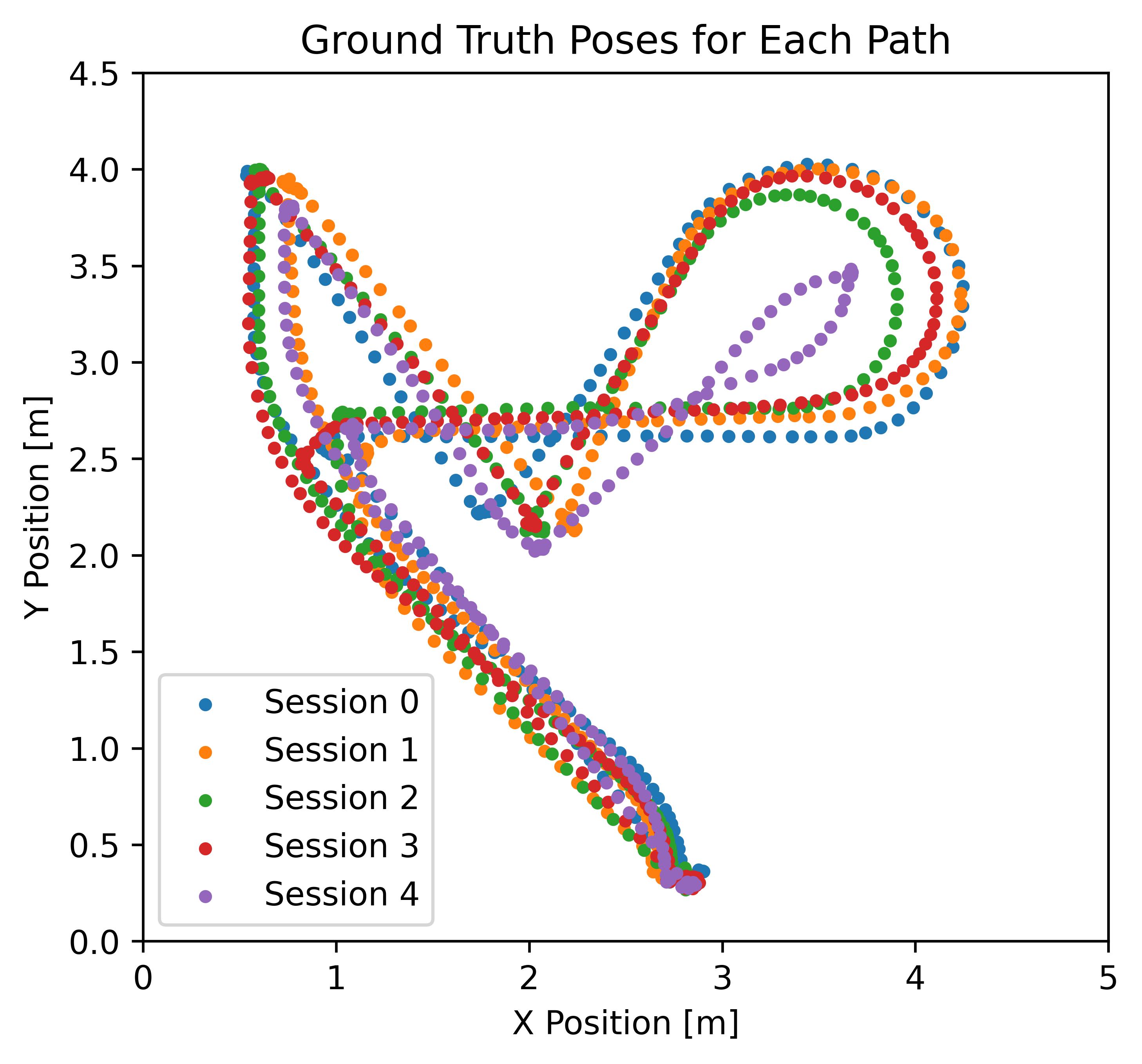}}
    \caption{\textit{Left:} Robot used for data collection, including a downward facing camera and motion capture markers for measuring ground truth. \textit{Right:} The ground truth pose of the robot for each camera observation across all sessions recorded.}
    \label{figure:robot-setup-and-ground-truth}
    \vspace{-7 mm}
\end{figure}

We configured a robot with a downward facing camera, measured and calibrated as described in Sec.~\ref{section:problem-description}. Fig.~\ref{figure:robot-setup-and-ground-truth} shows the setup. The camera is an Intel RealSense D435i located at a height of approximately 0.72 m above the ground. After rectifying the images, they are 1266x711 pixels. The robot starts each session in approximately the same location, then drives through a series of poses, collecting a color image of the ground at each pose. The path taken in each session is similar, but not identical, to ensure sufficient loop closures. Fig.~\ref{figure:robot-setup-and-ground-truth} shows these paths. Following the conventions established by the prior datasets~\cite{zhang_high-precision_2019},~\cite{schmid_hd_2022}, the robot stops at each pose to collect image and ground truth pose information. The operator manually assures each successive pose has sufficient overlap with the previous image to enable the visual odometry backbone. Qualitatively, this works out to approximately one quarter to one half of the image overlapping between successive frames. This intentionally precludes analysis of the trade space between system velocity, camera height, field of view, image overlap, and motion blur. In effect, this provides a ``best case" dataset for analysis. If a method does not work on this dataset, it is unlikely to work when undergoing motion. A discussion of the impact of these factors is included in Sec.\ref{section:pose-accuracy}.

Additionally, we used a Qualisys motion capture system and markers on the robot to record millimeter-level ground truth pose information for the robot at each observation. Using the markers on the robot, we defined a rigid body representing the robot with its origin directly on the ground, $+X$ axis facing out the front of the robot, and $+Y$ axis to the left of the robot. This allowed convenient mapping from 6DoF coordinates to 3DoF by extracting the X and Y coordinates of the robot's origin as well as the yaw. We verified that the Z coordinate, roll, and pitch were sufficiently close to zero to safely ignore. This ground truth information enables evaluating the accuracy of the proposed systems against an objective benchmark, rather than comparing against existing SLAM systems as is typical.

The entire data collection yields five sessions. Each session contains between 225 and 249 images and associated ground truth poses. Data is structured according to the format used by Schmid et al.~\cite{schmid_hd_2022}, modified slightly for poses in meters instead of pixels. The Appendix provides a link to the data.

\vspace{-2 mm}
\subsection{Quantitative Results}
\label{section:quantitative-results}
\vspace{-1 mm}
The primary quantitative metric of interest is overall accuracy of the pose estimates produced by each proposed system. Of secondary interest, but still highly relevant, is the average time to process new observations as they are received. The code to conduct the experiments described in this and the next section is linked in the Appendix.

\subsubsection{Pose Accuracy}
\label{section:pose-accuracy}
To measure pose accuracy of each system, the experiment proceeds as follows. The system is first instantiated. Then, observations from a single session are input to the system sequentially, which processes them according to the relevant method described in Sec.~\ref{section:multi-session-slam-methodologies} B-D. As SLAM maps are relative to an arbitrary map frame, the first measurement is assigned the known ground truth pose and a corresponding small covariance matrix. This facilitates evaluating accuracy against ground truth without loss of generality. At the end of each session, the estimated poses for that session are then recorded. This repeats for all sessions. After all poses have been estimated, we use the ground truth poses captured with the dataset to calculate the root mean square error (RMSE) for the position and orientation components of the poses.

Results for each proposed system are shown in Table~\ref{table:accuracy}. They are compared against several other systems. First is ORB-SLAM3~\cite{campos_orb-slam3_2021}, as it has multi-session monocular SLAM capability. Additionally there are several variations of the ground texture SLAM system proposed in Hart et al.~\cite{hart_monocular_2023} (which is summarized in Sec.~\ref{section:foundational-system} and forms the basis for the three methods under evaluation) as it is the only prior ground texture SLAM system. In the Original (Single) variant, all sessions are treated as one long single session. In the Original (Many) variant, each session is estimated with a brand-new instantiation of the system with no prior knowledge of previous sessions. Additionally, a variant of the original system that purely performs visual odometry with no loop closure detection and correction is also included.

\begin{table}
    \caption{RMSE metrics for the estimated pose versus ground truth for each system considered. \textit{Note:} ORB-SLAM3 was unable to produce continuous trajectories, so no error metrics could be calculated.}
    \centering
    \begin{tabular}{|c|c|c|}
        \hline
        \textbf{Method} & Position [\(m\)] & Orientation [\(deg\)] \\
        \hline
        KLD & 0.086 & 1.572 \\
        KLD (Grayscale) & \textbf{0.085} & \textbf{1.530} \\
        Visual Overlap & 0.212 & 8.153 \\
        Joint Intensity Histogram & 0.324 & 19.629 \\
        \hline
        ORB-SLAM3 & N/A* & N/A* \\
        Original (Single) & 0.540 & 27.013 \\
        Original (Many) & 0.266 & 14.963 \\
        Odometry & 0.210 & 7.244 \\
        \hline
    \end{tabular}
    \label{table:accuracy}
    \vspace{-6 mm}
\end{table}

\begin{figure}
    \centering
    \includegraphics[width=0.4\columnwidth]{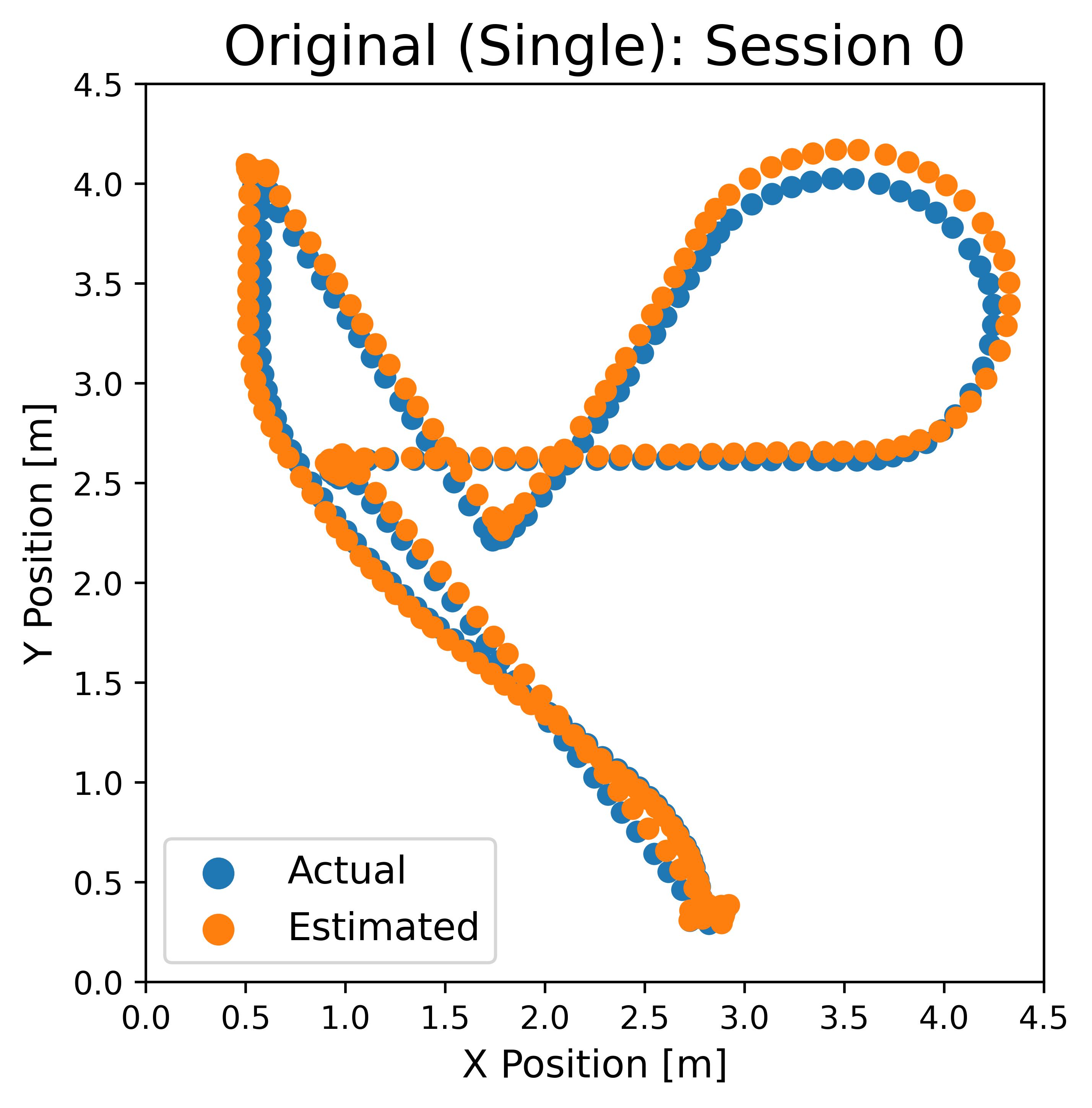}
    \includegraphics[width=0.4\columnwidth]{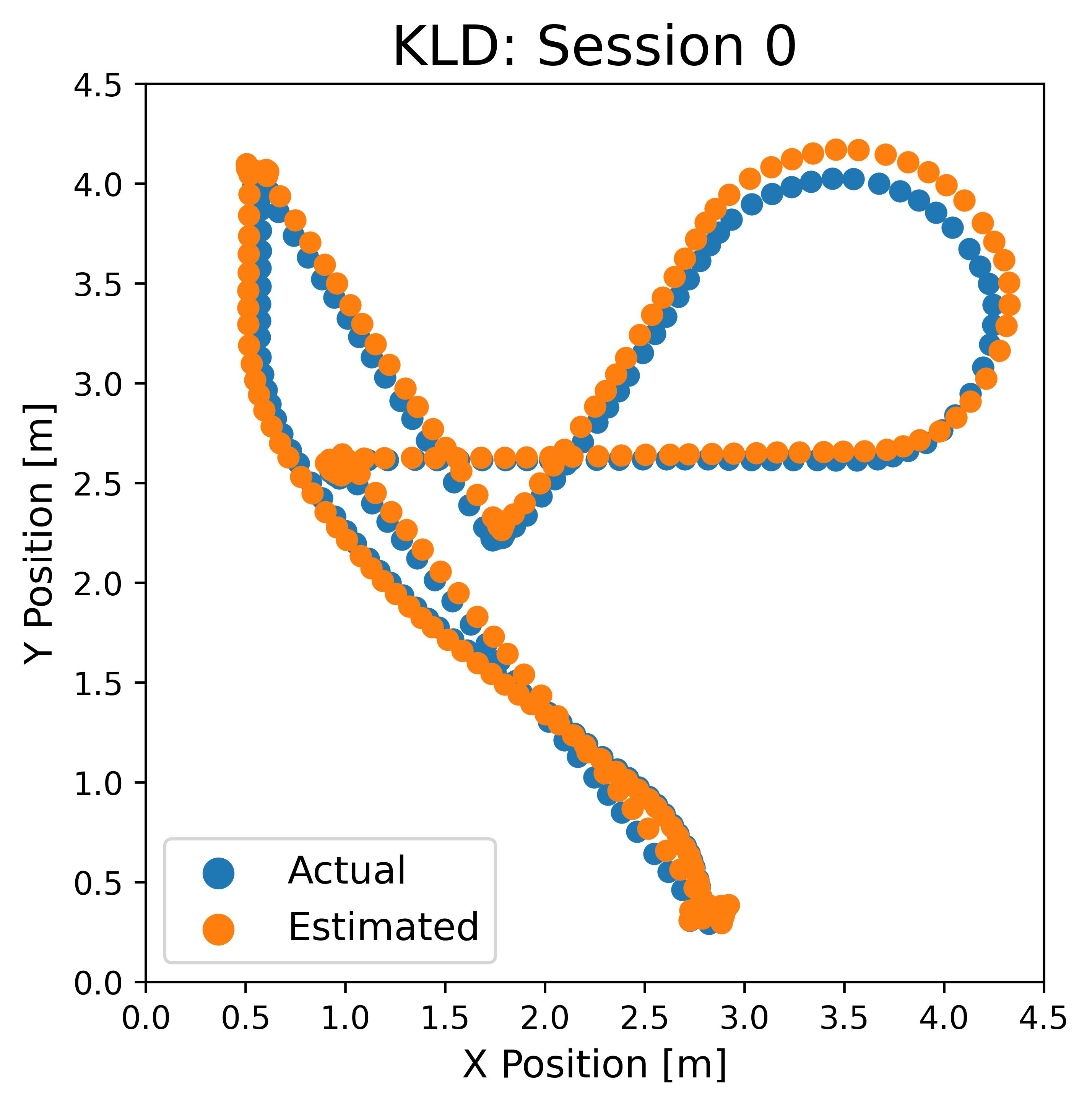}
    \\
    \includegraphics[width=0.4\columnwidth]{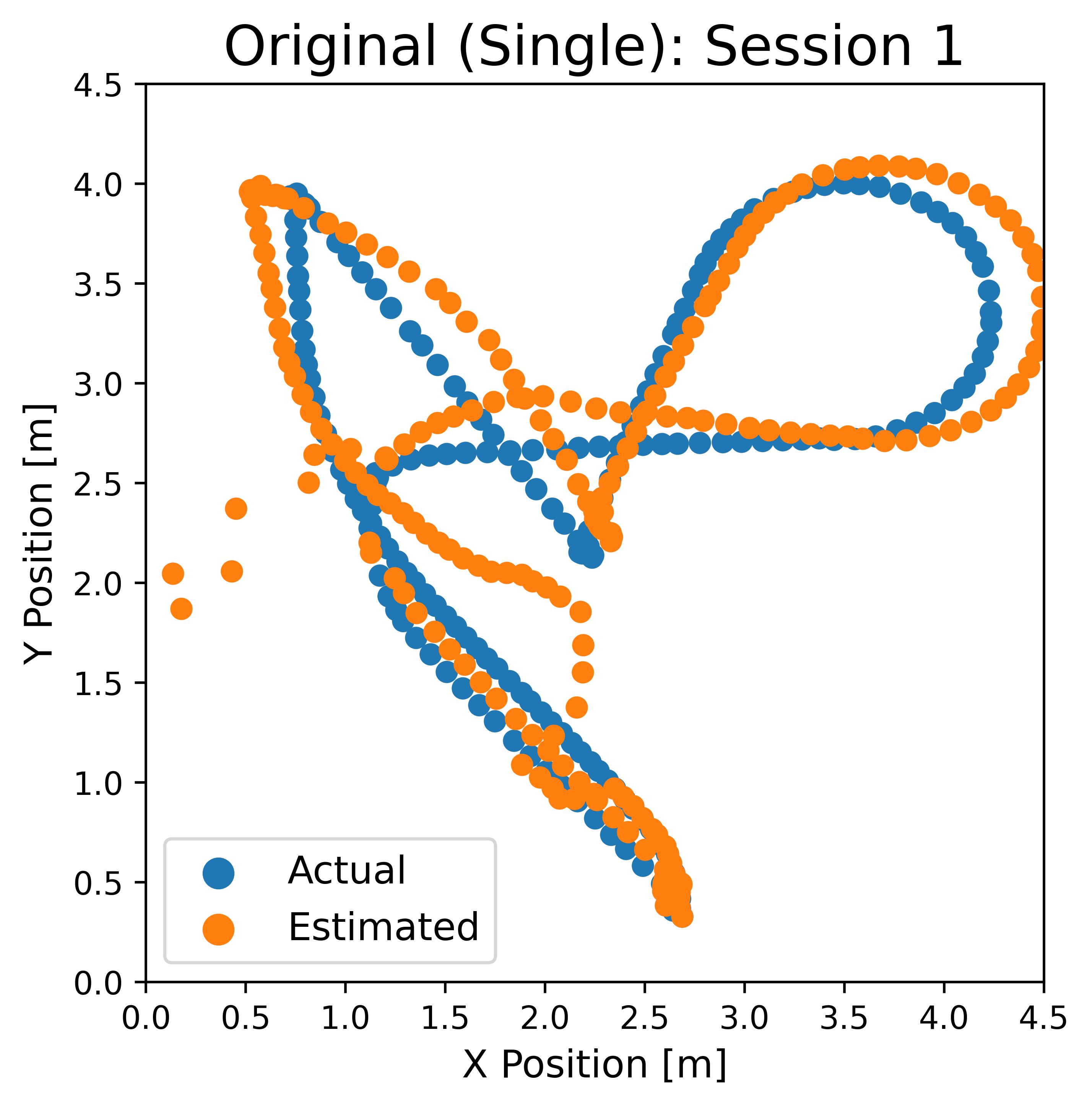}
    \includegraphics[width=0.4\columnwidth]{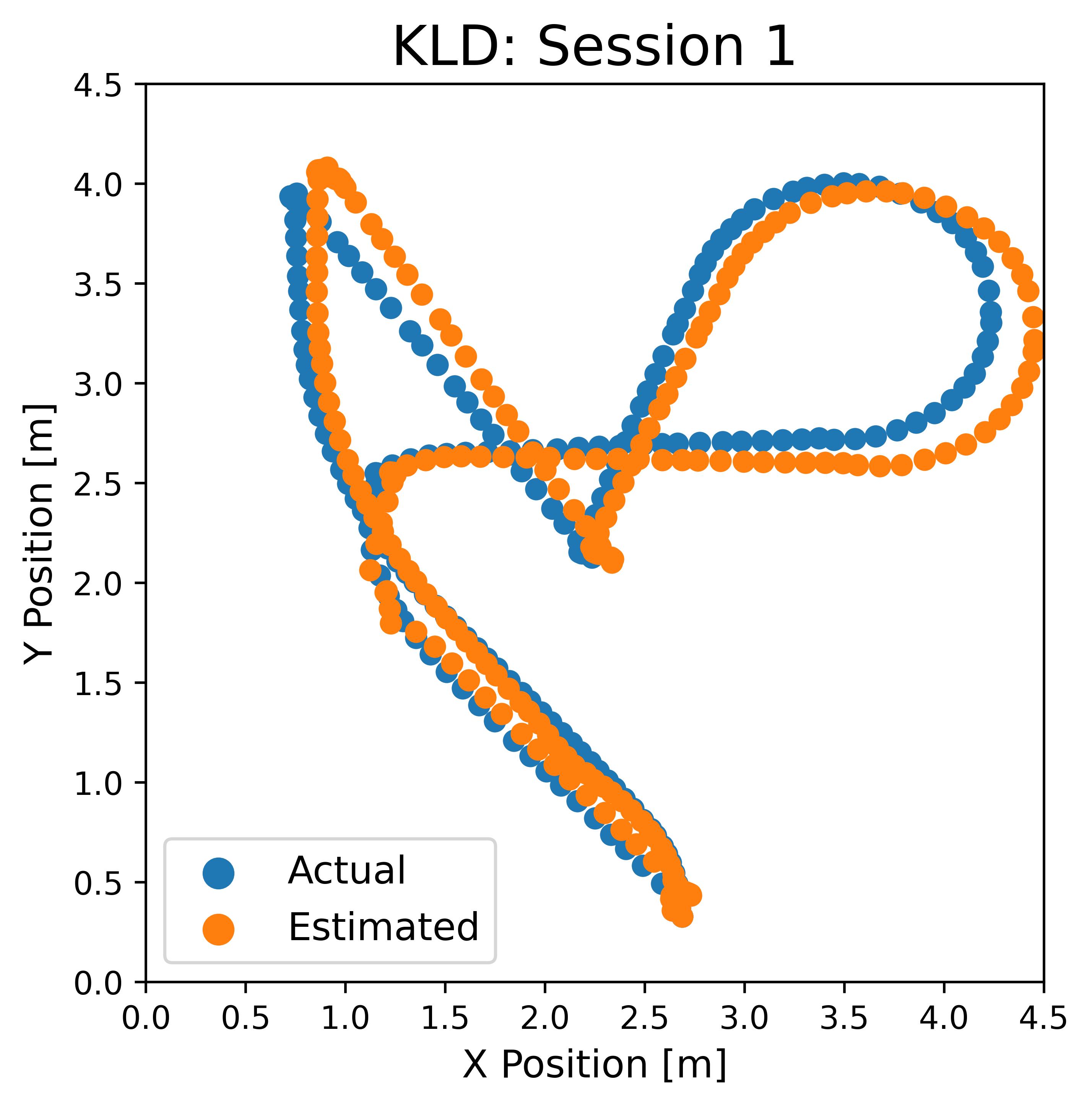}
    \\
    \includegraphics[width=0.4\columnwidth]{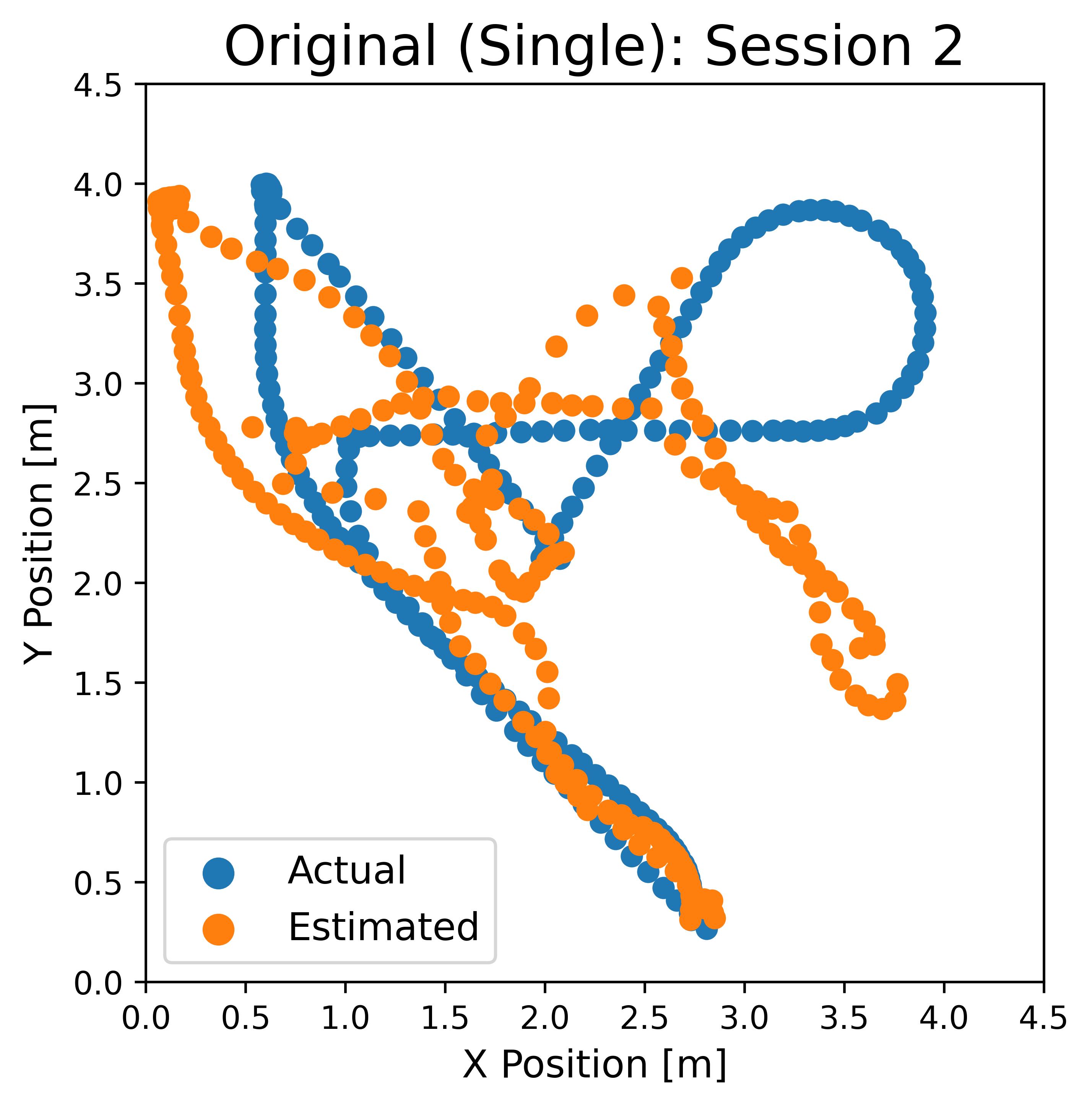}
    \includegraphics[width=0.4\columnwidth]{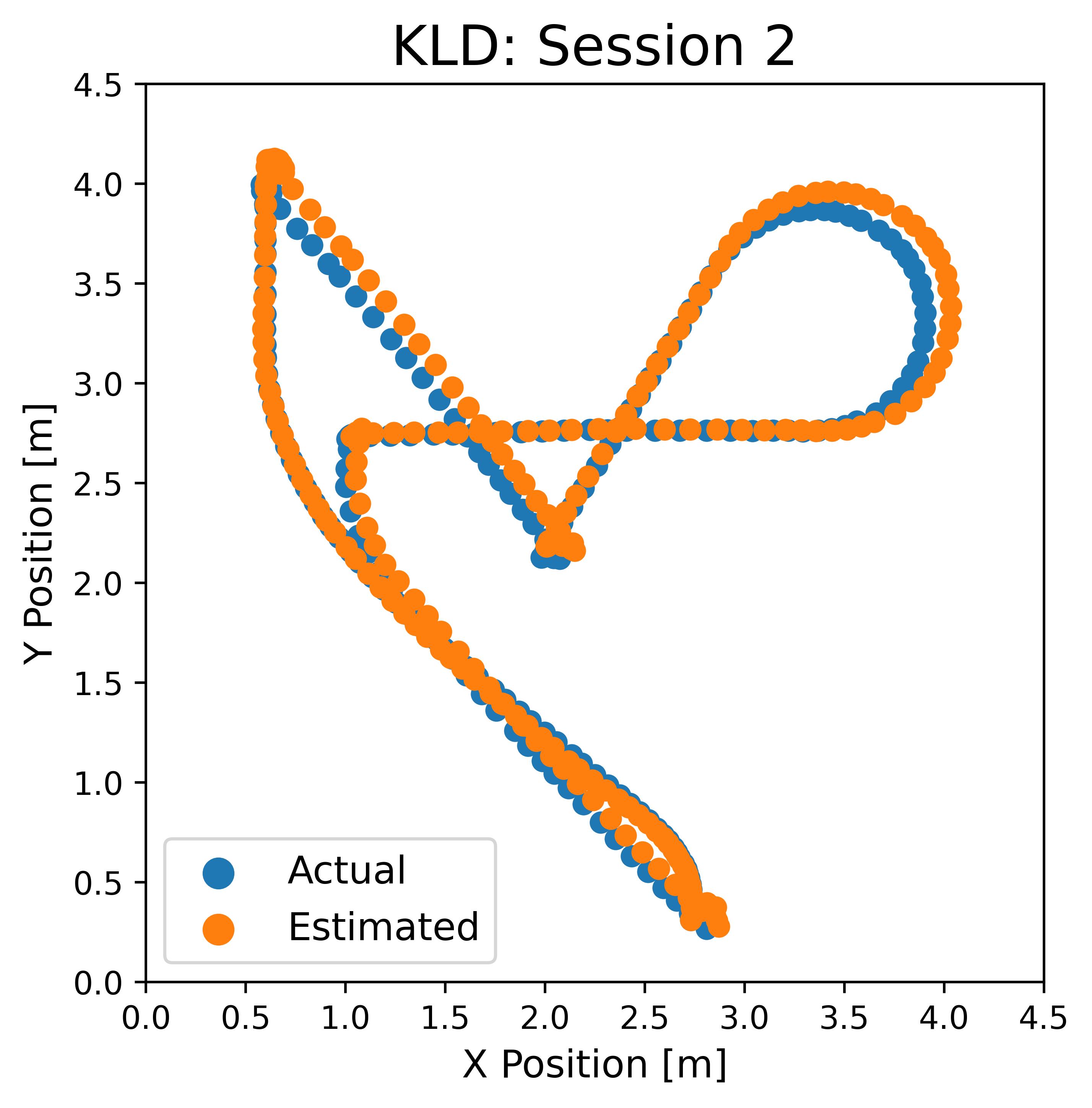}
    \\
    \includegraphics[width=0.4\columnwidth]{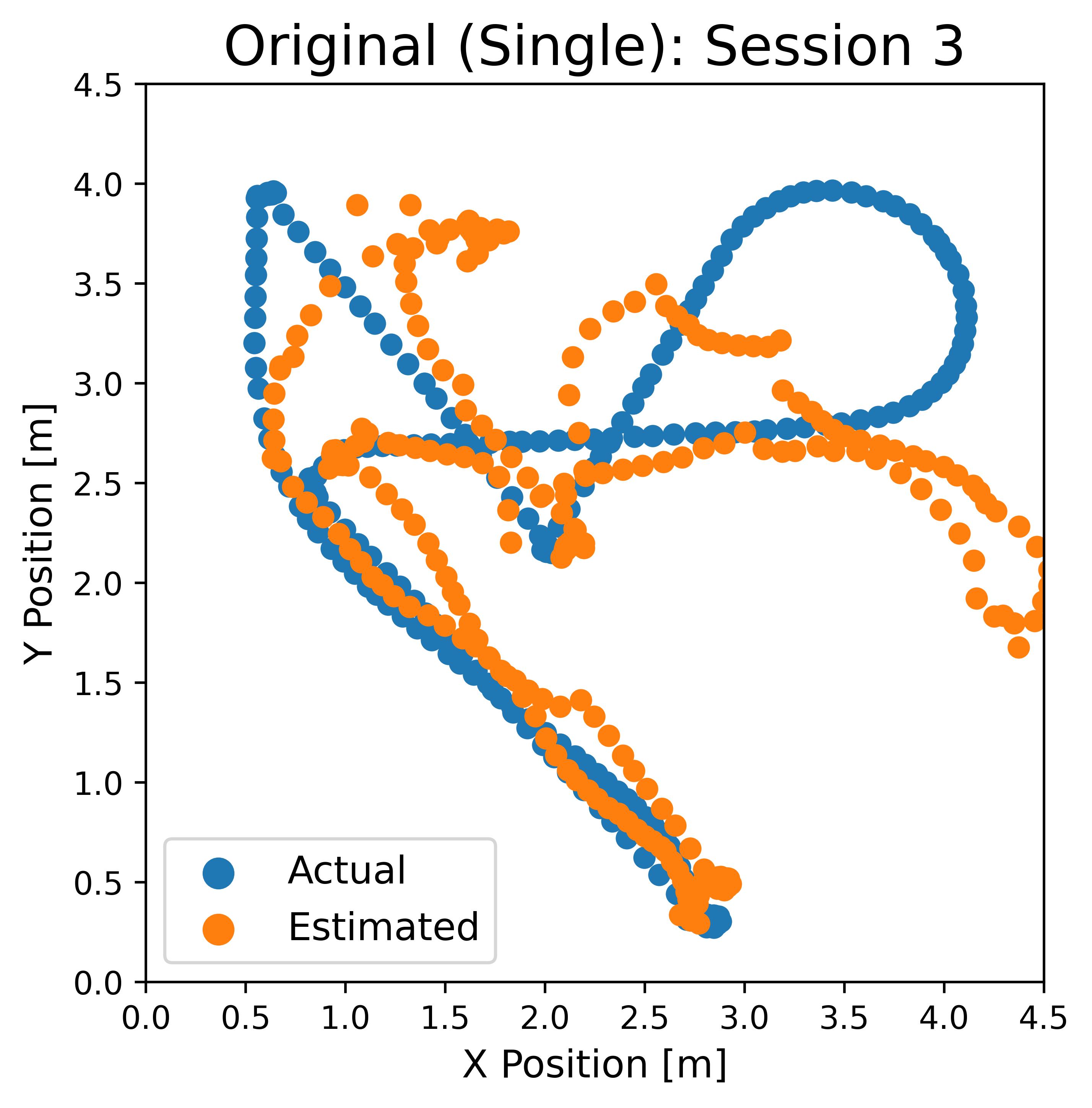}
    \includegraphics[width=0.4\columnwidth]{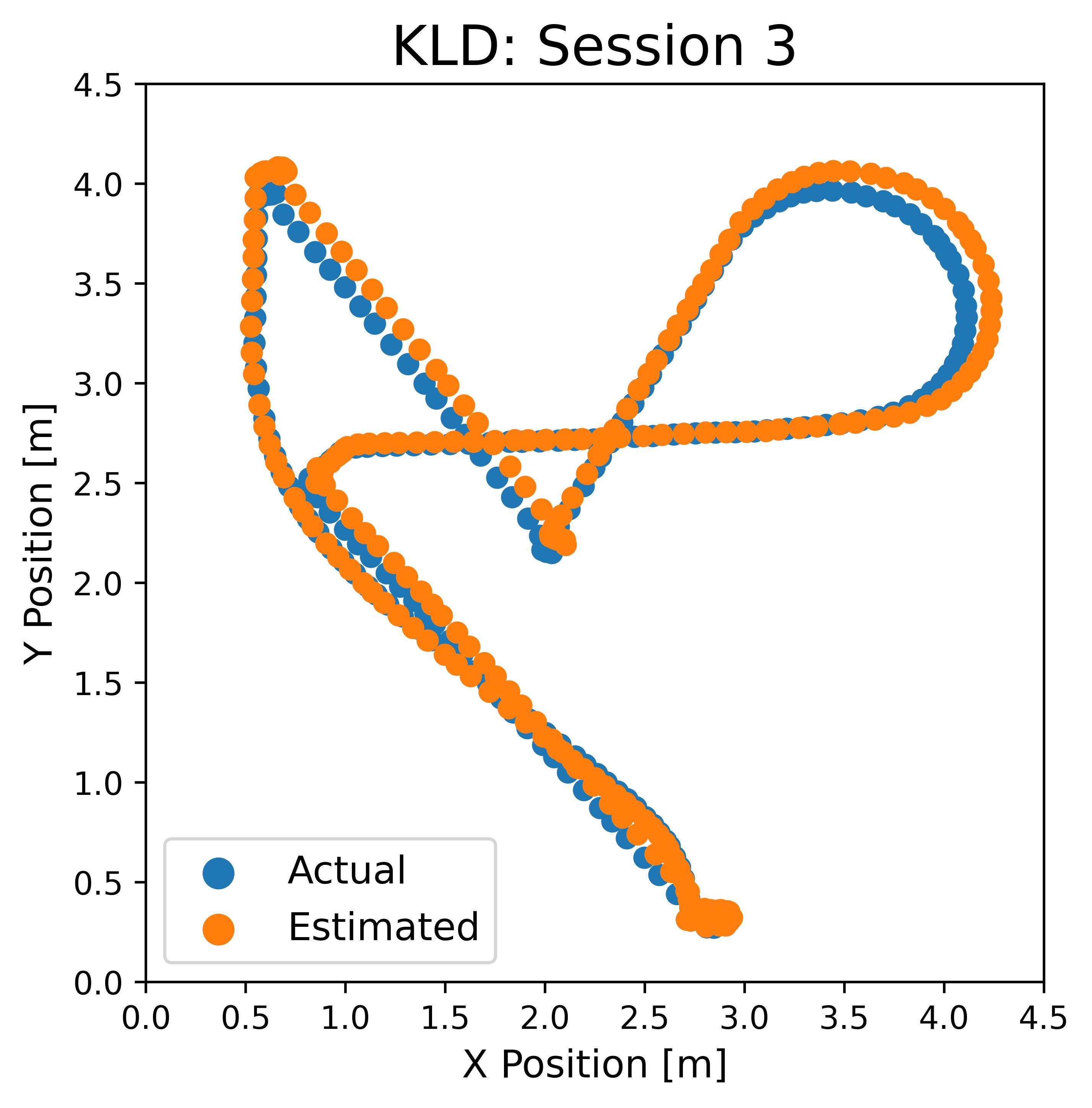}
    \\
    \includegraphics[width=0.4\columnwidth]{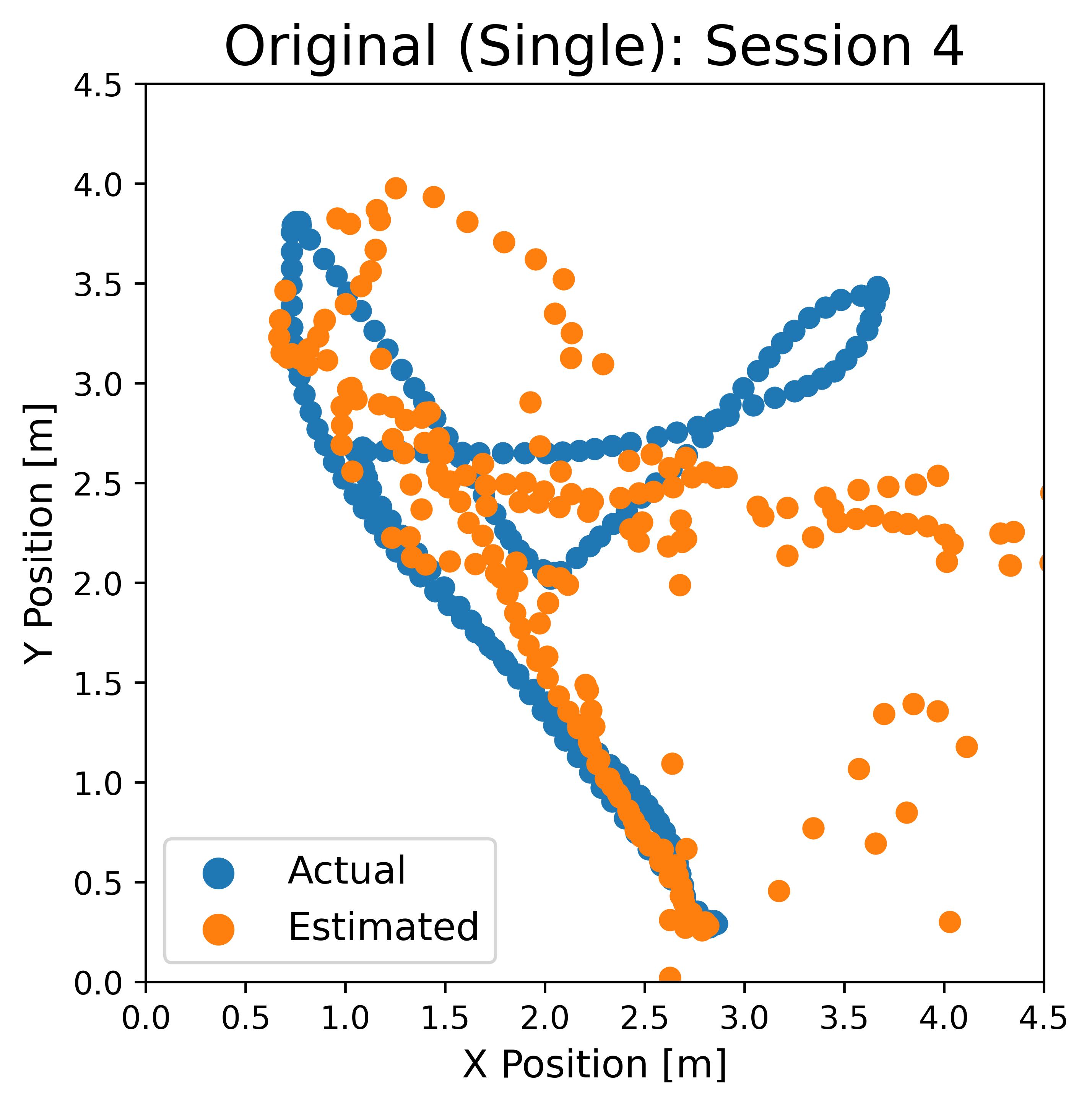}
    \includegraphics[width=0.4\columnwidth]{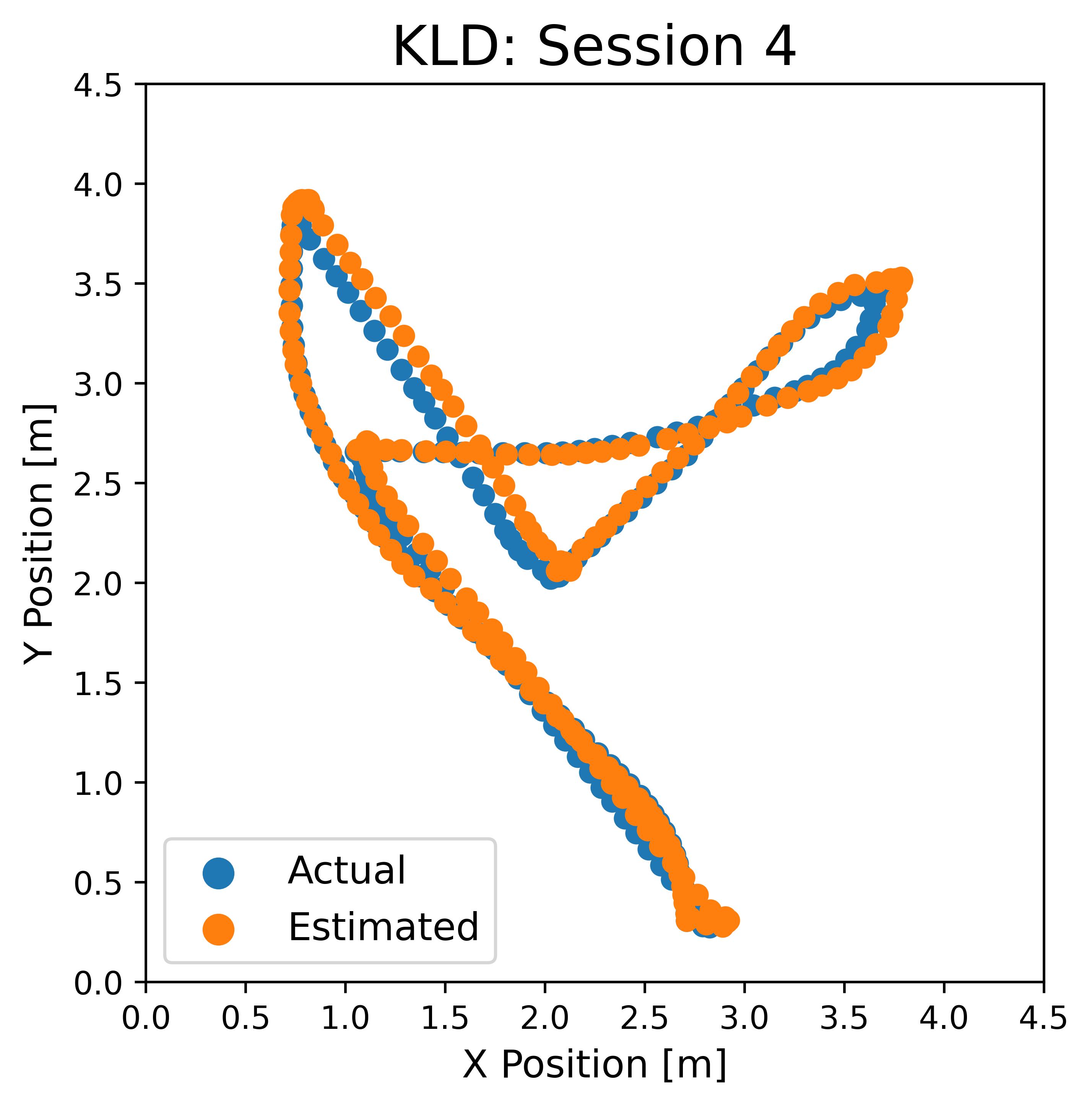}
    \caption{The estimated poses for each session using the Original (Single) SLAM system (\textit{left}) and the KLD system (\textit{right}).}
    \label{figure:session-plots}
    \vspace{-7 mm}
\end{figure}

As can be clearly seen, the KLD method offers significantly better results over any of the other methods, both when using the three-channel color version or single-channel grayscale version. This can also be seen in Fig.~\ref{figure:session-plots}, which shows the estimated poses using the KLD system and the Original (Single) system compared to ground truth. While both systems perform equally well in the first session, Original (Single) quickly diverges as the texture starts to change, whereas the KLD version remains accurate across all sessions. In most cases, the failure of the Original (Single) results from erroneous loop closures distorting the factor graph. Introduction of KLD avoids these issues.

% The inkscapepath puts the outputs in different subdirectories so that the image names do not cause conflicts:
% https://tex.stackexchange.com/questions/707280/includesvg-for-files-with-the-same-name-in-different-folders-directories

The other systems are not shown, but their performance is typically equally poor, with very large errors and misalignments in the later sessions. In the case of ORB-SLAM3, despite being the standard benchmark for forward-looking SLAM systems, the system failed to produce any consistent trajectory, regardless of the parameters used. Potential causes include visual similarity of the data proving challenging for the place recognition component, and/or the lack of near and far features for transform estimation. In either case, this result illustrates the advantages of a domain-specific system when working with ground texture.

Not included in this analysis is the impact of camera height and robot velocity. Were the camera too close to the ground, it would be less likely for any two frames to share overlapping fields of view. This limits the amount of loop closures the system can identify and the accuracy of the visual odometry estimate. Both degrade system performance. Inversely, if the camera were higher, there would be a greater number of overlapping fields of view at the risk of insufficient image quality to identity distinct features in the images. However, a higher resolution camera mitigates this.

Likewise, the velocity of the robot has similar impact. A slow velocity would produce a larger number of overlapping images, likely yielding better performance. Too fast of a velocity would degrade the visual odometry estimate, as successive images will have less overlap. The loop closure identification process could correct some of this, but overall accuracy is expected to suffer. Additionally, motion blur would reduce the number of identifiable distinct features used for matching. It generally is not expected to impact KLD scores though, as they are based on overall pixel color values in the image and not specific features. A faster camera frame rate and shutter speed would mitigate these impacts.

\subsubsection{Timing}
\label{section:timing}
Understanding the average time for the system to process new measurements provides insight into the scalability of the system. In particular, it is important to understand the impact of the new multi-session components (i.e., KLD) over and above the previous system.

\begin{figure}[ht]
    \centering
    \includegraphics[width=0.9\linewidth]{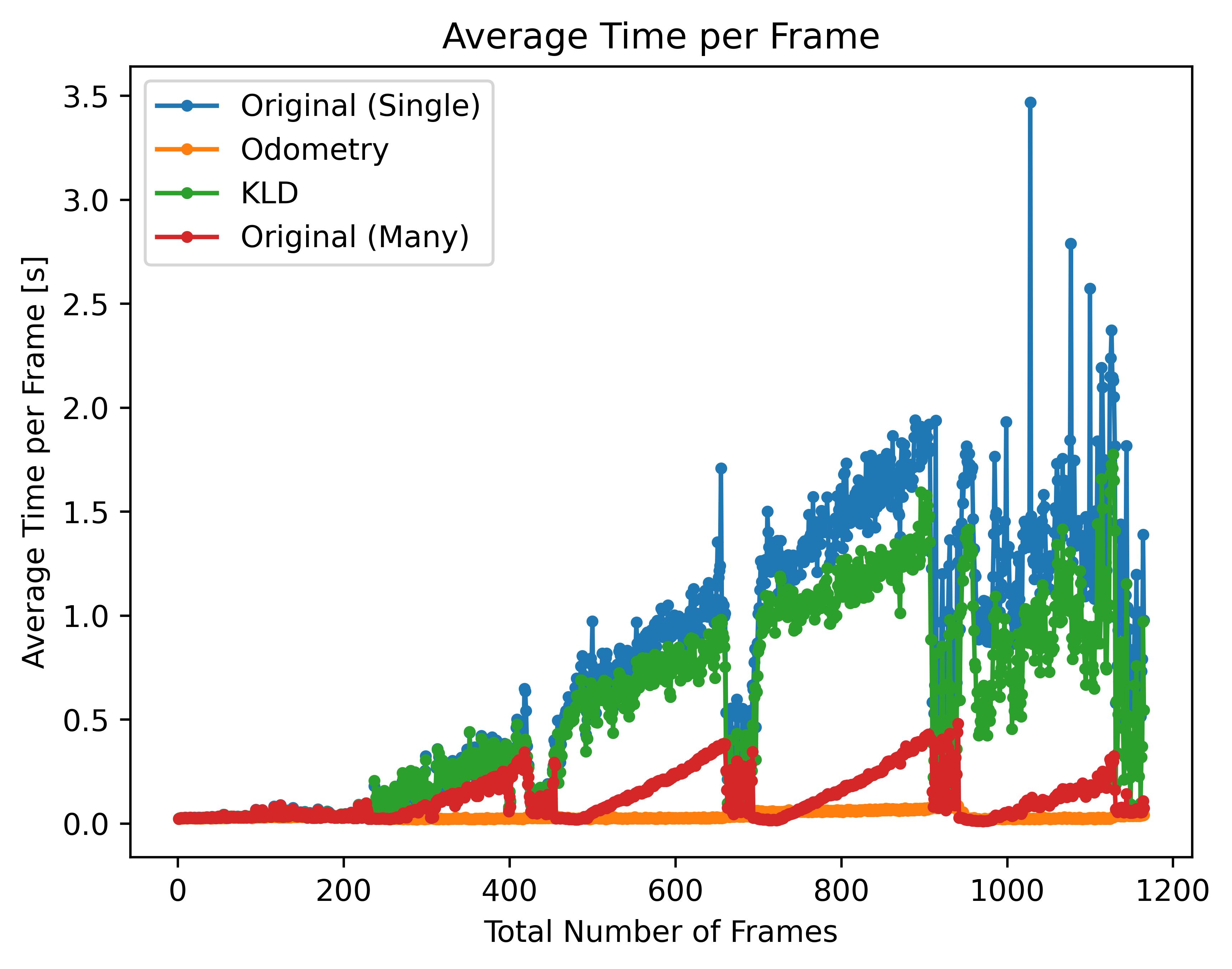}
    \caption{The average time for KLD and several variants of the original system to process a particular measurement.}
    \label{figure:timing-plot}
    % Make just a bit smaller so that it visually looks the same.
    \vspace{-3 mm}
\end{figure}

To measure this, we initialize ten instances of each system. Then, for each observation, we add it to each instance of the system and wait for that instance to finish processing before adding the observation to the next instance of the system. We record the total time across all ten instances and find the average time to process that observation. We then proceed to the next observation and continue this approach until all observations have been added. Any processing performed at the end of each session is not included, as these steps do not need to be performed as the system operates. The result is a rough timing profile, which is shown in Fig.~\ref{figure:timing-plot}. The only multi-session method shown is KLD, for clarity and since it is the most accurate version. As with the pose accuracy, it is compared against three variants of the original SLAM system. ORB-SLAM3 is excluded due to its issues producing a valid trajectory. These measurements were captured on a Dell Precision 7780 with an i9 processor and 64 GB of memory. When running, as many other processes as possible were stopped, but some background tasks were still running.

\vspace{-3 mm}

As can be seen, the KLD system adds no additional computation overhead compared to the Original (Single) SLAM version. This matches a computational complexity analysis. The KLD method described in Sec.~\ref{section:kullback-leibler-divergence} only performs two operations for each measurement: the computation of a histogram and the computation of a KLD score. The KLD scoring is \(O(1)\) as it is just two calculations. Computing the current image histogram scales with the number of image pixels, but is still \(O(1)\) with respect to the number of observations stored in the map, as it only uses the most recently received observation. Computing the baseline histogram can be done offline after the first session, so is not counted. The result is an approximately constant, small additional computation time added to each observation processing step. When combined with variability of processing time due to other computer processes, approximately equal time for both the KLD system and Original (Single) as seen here is reasonable. This result is highly beneficial for multi-session systems, as it scales no worse than the original system, but is more robust to change.

It should be noted that scalability of the original SLAM system is not under consideration. While a linear scaling may not be sufficient for a long-running SLAM system, the important conclusion is that the addition of a KLD method to adapt for multi-session systems does not worsen the computational time-scaling. Any subsequent improvements in overall system performance in the underlying system will benefit the multi-session version as well.

The key finding for system timing is that the system experiences only a very minimal, constant-time performance hit, at the advantage of gaining increased robustness to low-dynamic changes in the ground texture of the environment.

\vspace{-2 mm}
\subsection{Detailed Examination of KLD Approach}
\label{section:detailed-results}
\vspace{-1 mm}
Now that the KLD method is found to be the most beneficial, further analysis can illustrate the impact this approach has on overall performance. Additionally, there is a tangential benefit for environment maintainers that is important to highlight.

\subsubsection{KLD's Impact}
\label{section:klds-impact}
To understand the impact of using KLD scores to adjust the pose graph, a deeper analysis of how it impacts the loop closure evaluation process is helpful.

First, we present an examination of how KLD scores change which potential loop closures are added to a multi-session SLAM solution. As ground truth information is available in the dataset, it is possible to determine if any two observations have overlapping fields of view. If they do, they are considered a true loop closure. Likewise, predicted loop closure labels can be found by logging diagnostic data within the appropriate system.

\begin{figure}
    \centering
    \includegraphics[width=0.49\linewidth]{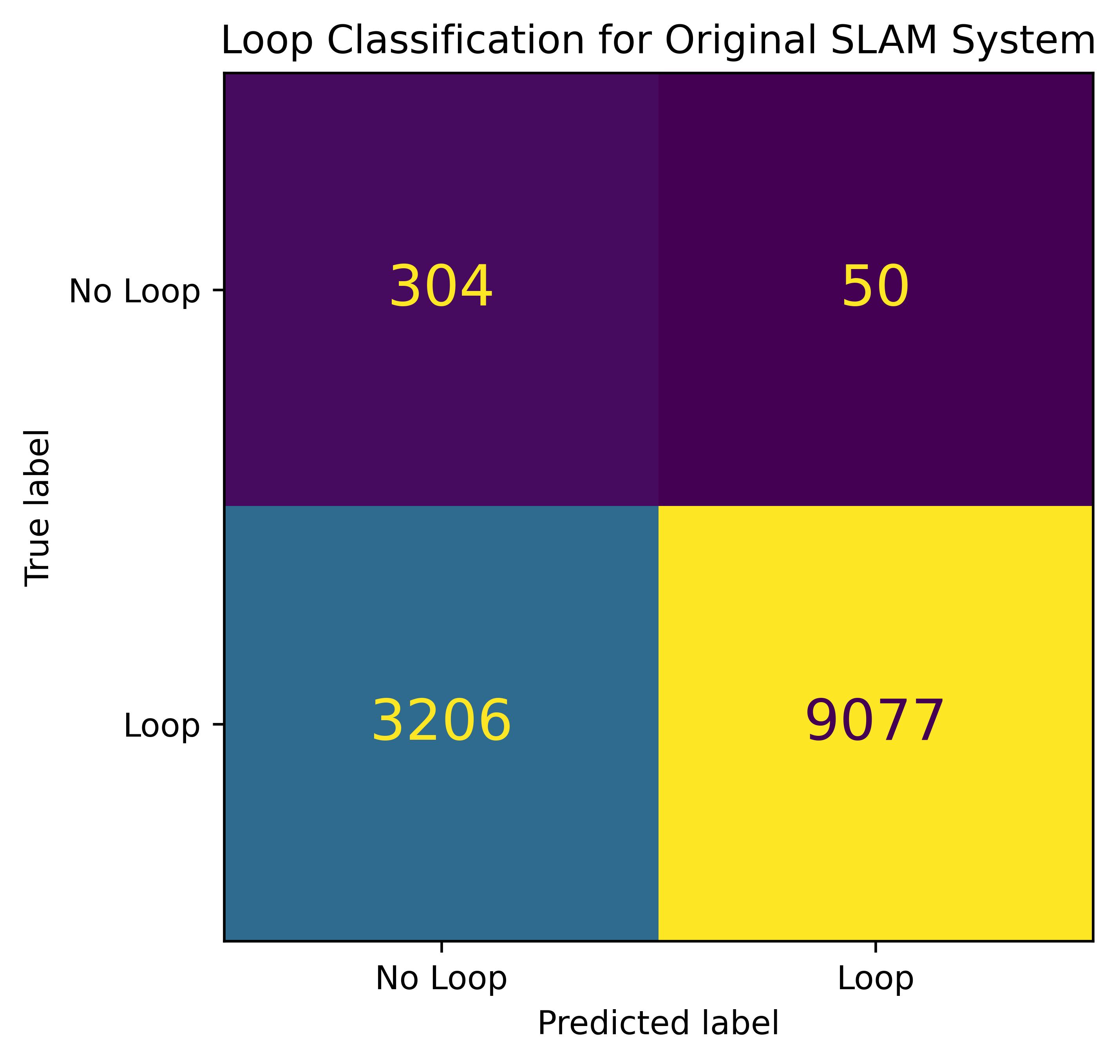}
    \includegraphics[width=0.49\linewidth]{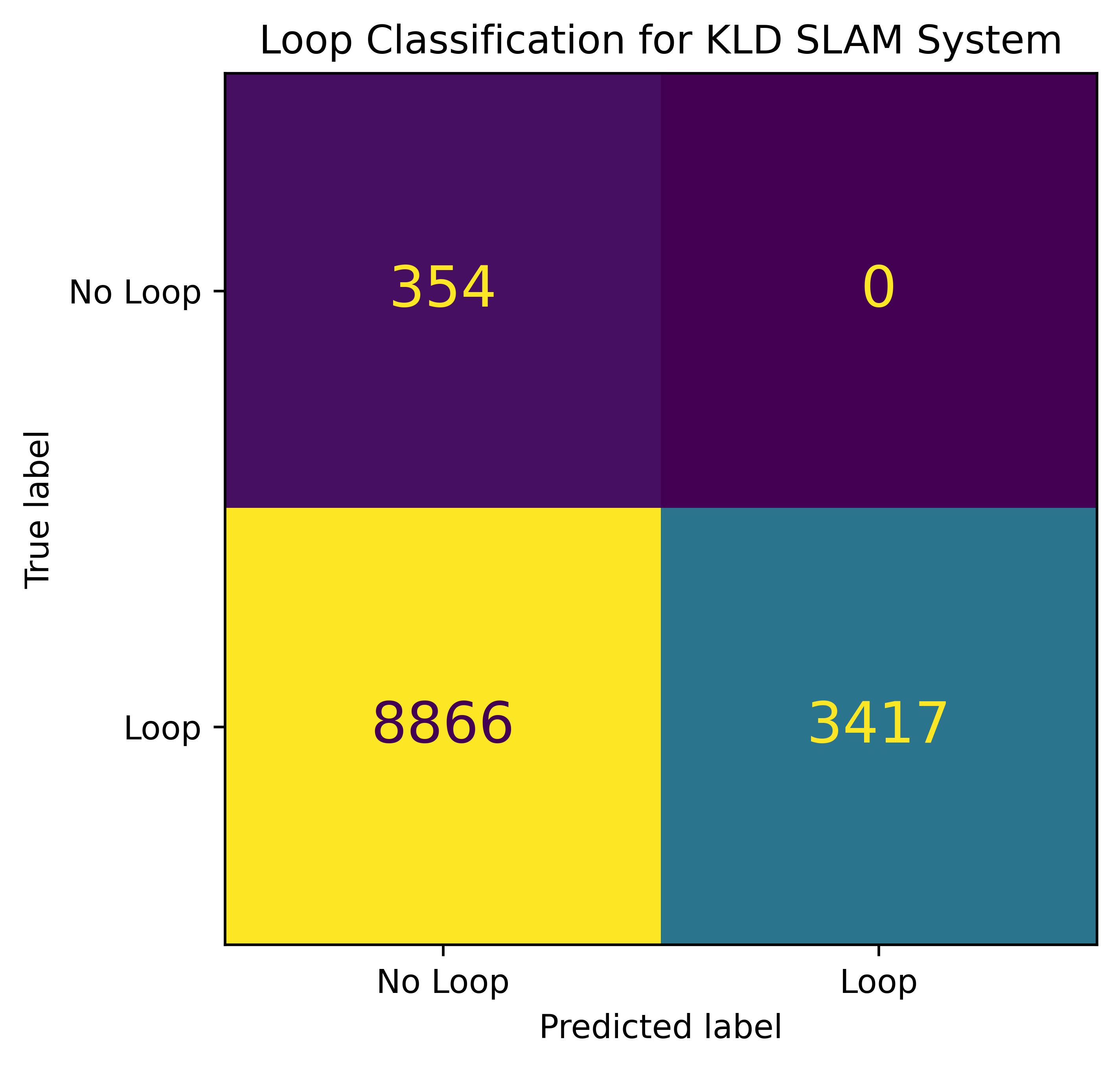}
    \caption{A comparison of how many loop closure candidates the systems add to the pose graph or not, versus if the candidate is actually a loop closure based on ground truth data. The Original (Single) SLAM system is at left and the KLD system is at right.}
    \label{figure:confusion-matrix}
    \vspace{-7 mm}
\end{figure}

Fig.~\ref{figure:confusion-matrix} shows a breakdown of how loop closure candidates are classified by each version of the system. Of particular note is the absence of false positives in the KLD version, meaning no erroneous loop closures are added to the pose graph. As compared to the 50 erroneous loop closures added by the Original (Single) SLAM system, this represents a noticeable improvement in accuracy.

\begin{figure}[t]
    \centering
    \includegraphics[width=0.75\linewidth]{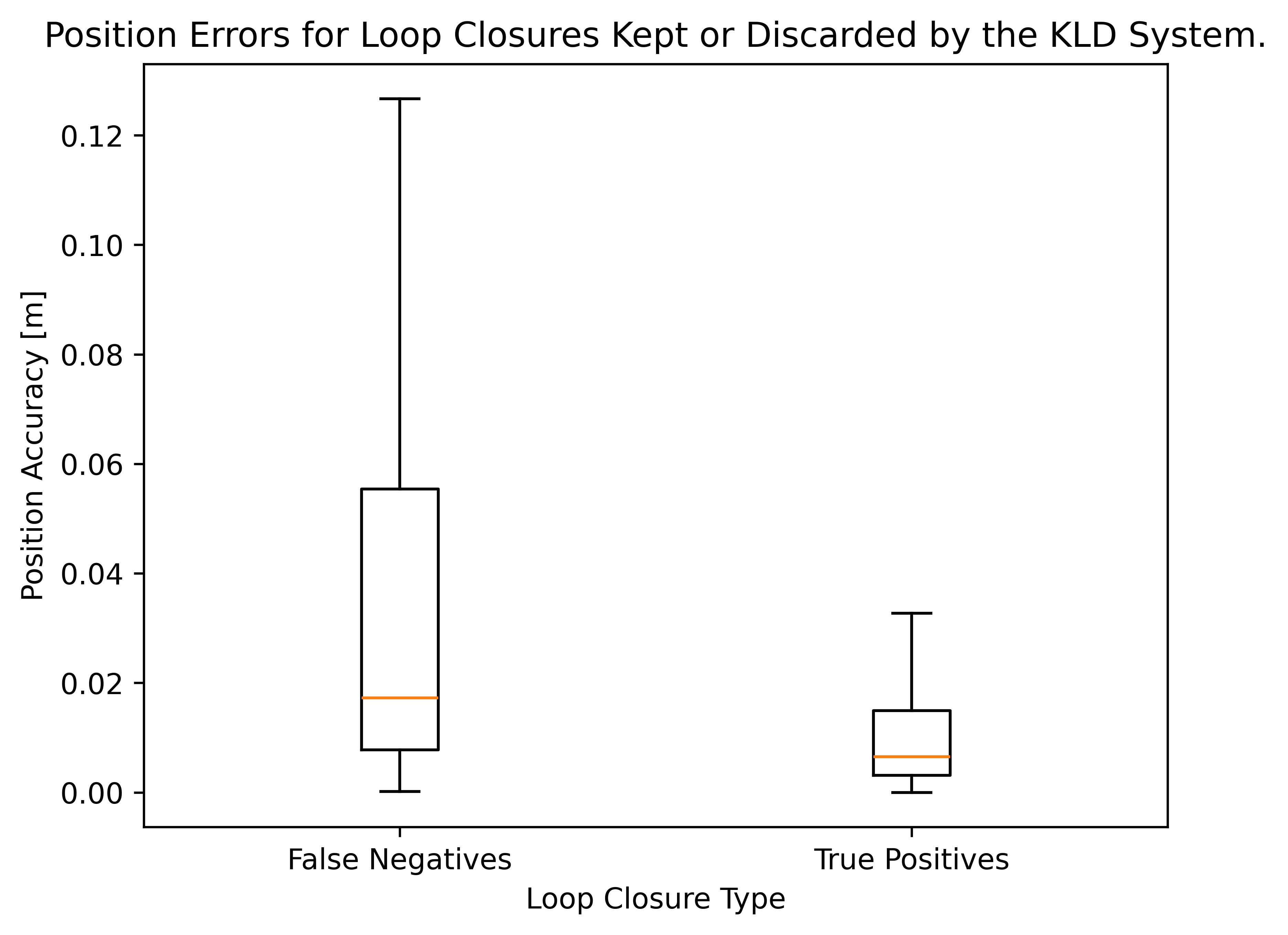}
    \caption{The position accuracy of true loop closure candidates removed (\textit{left}) and kept (\textit{right}) by the KLD system. Note that outliers are excluded for legibility.}
    \label{figure:position-error-box-plot}
    \vspace{-4 mm}
\end{figure}

Additionally, while Fig.~\ref{figure:confusion-matrix} does show some false negatives (i.e. loop closures incorrectly excluded from the pose graph) for the KLD system, the ones that are excluded have worse accuracy on average. Fig.~\ref{figure:position-error-box-plot} shows a comparison of position error between correct loop closures that the KLD system added to the pose graph and correct loop closures it excluded. The ones that are added have a higher accuracy overall. So even though some loop closures were incorrectly excluded, the overall quality of the map improved, leading to the accuracy seen in Sec.~\ref{section:quantitative-results}.

\begin{figure}[t]
    \centering
    \includegraphics[width=0.6\linewidth]{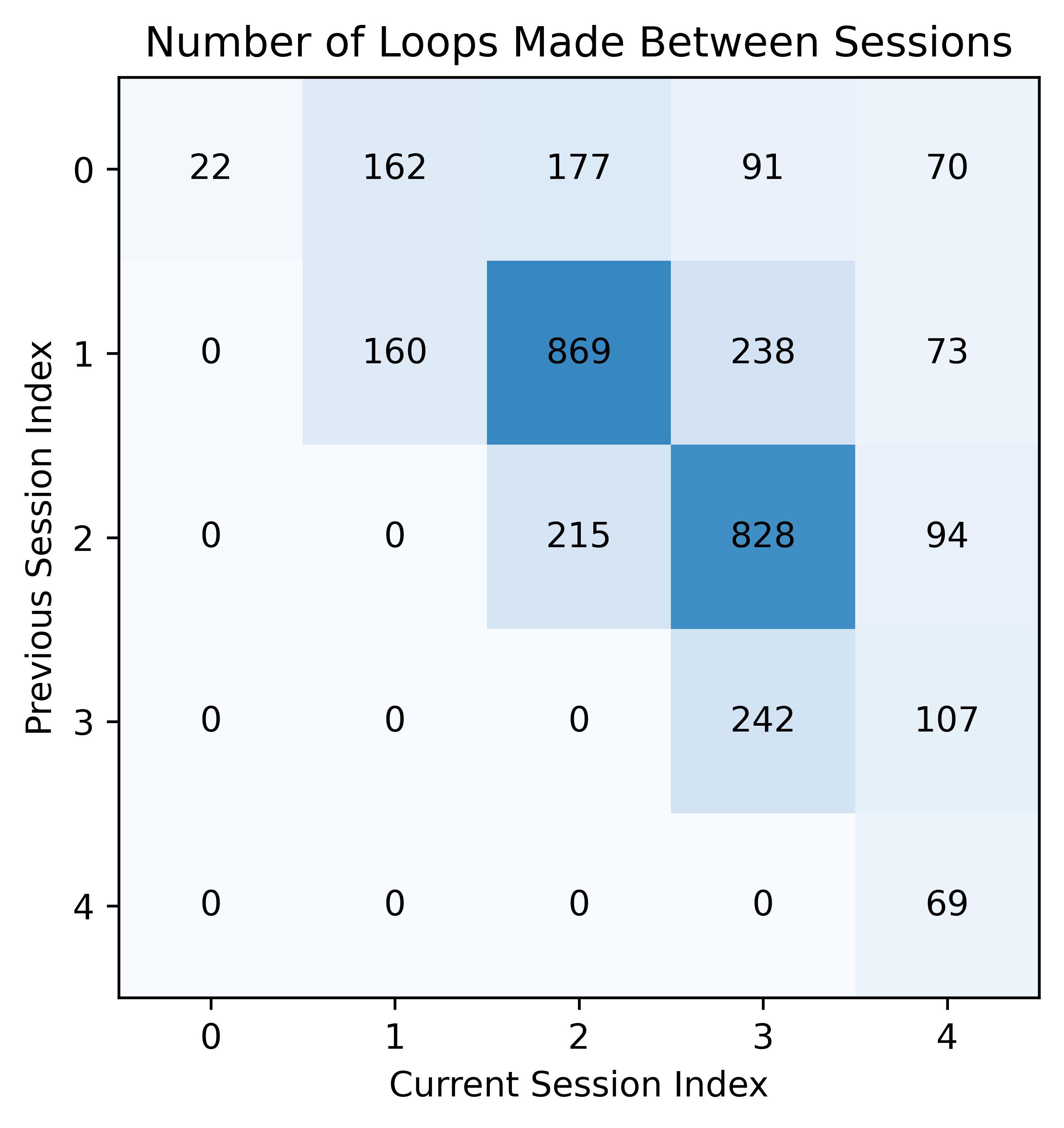}
    \caption{A mapping of each session involved in loop closures across the five total sessions in our dataset. The current session is the \(k\)th session that contains observation \(\mathbf{z}_{k,t}\) when it is added to the map. The previous session is the one containing the already existing observation to which the loop closure is made.}
    \label{figure:loop-closure-matrix}
    \vspace{-7 mm}
\end{figure}

To further illustrate the impact of a KLD score, Fig.~\ref{figure:loop-closure-matrix} shows the total count of which sessions contain the two observations that are related by each included loop closure. Notably, the farther apart the sessions are, fewer loop closures are added to the pose graph. In the dataset used here, each subsequent session contains a higher severity of change. Therefore, decreasing total loop closures as the distance between indices increases is consistent with a system that more often excludes loop closures subject to large-scale low-dynamic environmental changes, where correct transform estimation is unlikely.

Taken all together, this indicates a multi-session system that is well-situated to handle low-dynamic changes between sessions. It is capable of excluding incorrect loop closure candidates and ensuring that the loop closures which are added to the pose graph all are reliably likely to contain correct transform estimations.

\subsubsection{Wear Heat Map}
\label{section:wear-heat-map}
\begin{figure}[t]
    \centering
    \includegraphics[width=0.49\linewidth]{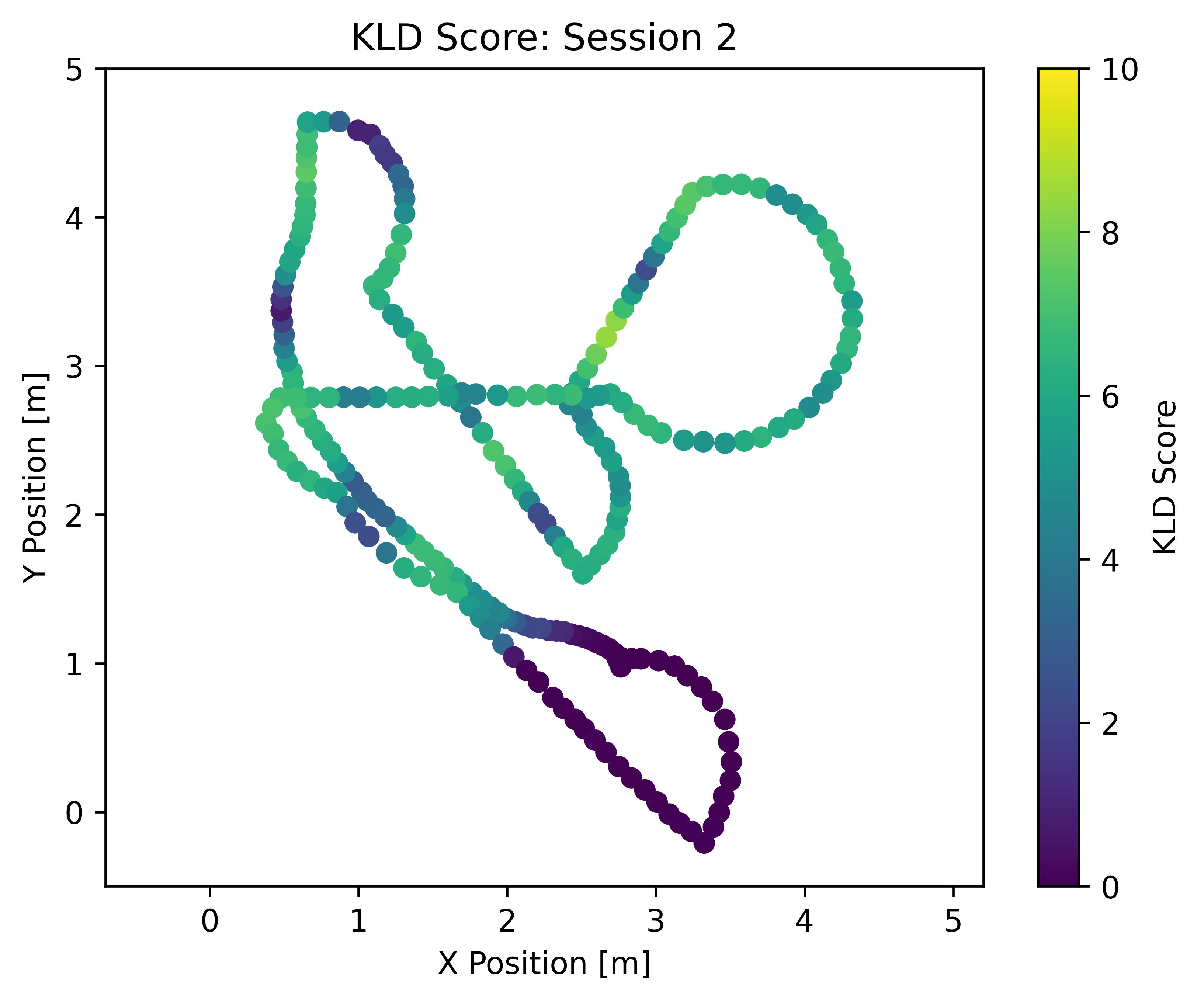}
    \includegraphics[width=0.49\linewidth]{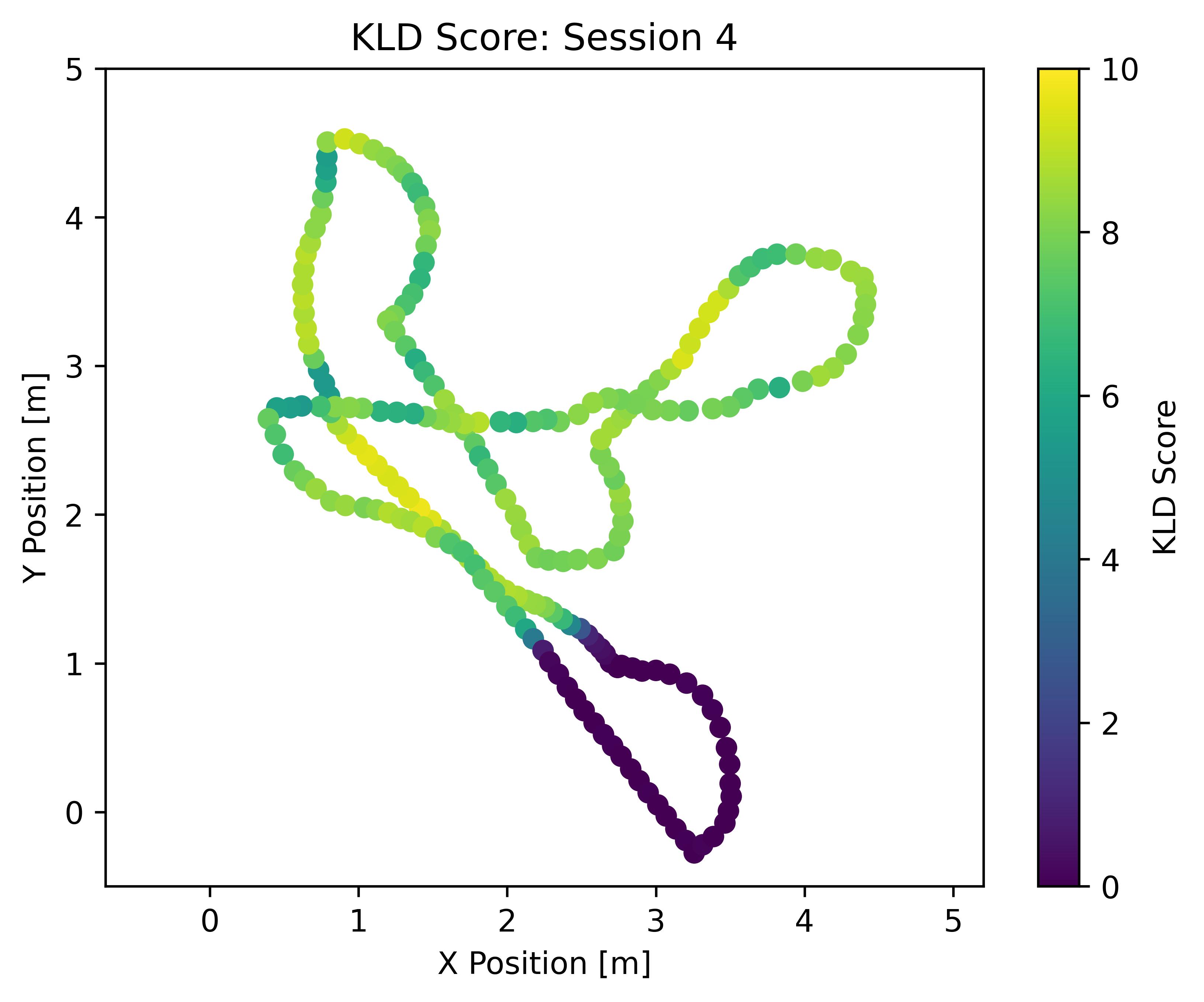}
    \caption{Texture wear heat maps generated using the KLD score and estimated position of the center of the image. Higher numbers and associated colors indicate more wear. Note that because this uses the center of the image, not the center of the robot, the estimated positions are different from previous figures.}
    \label{figure:health-map}
    \vspace{-7 mm}
\end{figure}

Evaluation of the KLD score for each image also provides benefits to maintainers of the operational environment as a signal that the surface is worn enough to possibly require repair. The score provides a numeric quantification of the difference between a given image and the baseline. When combined with an accurate estimate of the robot's pose when the image was taken, a heat map of surface wear can be easily constructed. Fig.~\ref{figure:health-map} provides an example for two of the later sessions. To construct this map, the center point of the image was projected into the world frame using the estimated pose of the robot at that time. The KLD score is then used to assign the color value.

Noticeably, the scores increase significantly in the regions of the environment that experience the biggest changes. While the exact numeric values may not have significance to the maintainers of an environment, this does provide a way to quickly identify potential spots of significant wear. With this information, maintainers can rapidly prioritize inspection and maintenance.
\vspace{-2 mm}

\section{Conclusions}
\label{section:conclusions}
\vspace{-1 mm}
We have highlighted the need for ground texture SLAM systems that are robust to low-dynamic environmental changes in order to deploy robots in the relevant operational environments for long durations. To address this need, we subsequently introduced and measured the effectiveness of three different methods on multi-session low-dynamic change ground texture data. We also introduced a dataset featuring both low-dynamic change ground texture and high-accuracy ground truth pose information to assist with measuring effectiveness. The accuracy evaluations show a clear winner: use of Kullback-Leibler Divergence to bias pose graph factor confidence. Our subsequent analysis then provided insight into this method's impact on timing, as well as showcasing its effectiveness at removing erroneous or poor accuracy loop closure candidates. We also briefly highlighted its usefulness as an environmental health monitor measure for maintainers. For any robot system operating for long durations in ground texture environments, this analysis and toolset provides a helpful means to ensure a robot's map of the environment stays correct throughout its entire lifespan.
\vspace{-1 mm}

\section*{Appendix}
\label{section:appendix}
% Make the URLs look nice.
The dataset and code described here are available at \\
https://gitlab.com/riselab/multi-session-ground-texture-data \\
and \\
https://gitlab.com/riselab/multi-session-ground-texture-slam.

\bibliographystyle{IEEEtran}
\bibliography{IEEEabrv,references}

\end{document}